\documentclass[journal]{IEEEtran}

\ifCLASSINFOpdf

\else

\fi

\usepackage{cite}
\usepackage{graphicx}
\usepackage{subfigure}
\usepackage{amsmath}
\usepackage{amssymb}
\usepackage{url}
\usepackage{algorithm}
\usepackage{algorithmic}
\usepackage{multirow}
\begin{document}

\title{StarCraft Micromanagement with Reinforcement Learning and Curriculum Transfer Learning}

\author{Kun~Shao,
        Yuanheng~Zhu,~\IEEEmembership{Member,~IEEE}
        and~Dongbin~Zhao,~\IEEEmembership{Senior Member,~IEEE}
\thanks{K. Shao, Y. Zhu and D. Zhao are with the State Key Laboratory of Management and Control for Complex Systems, Institute of Automation, Chinese Academy of Sciences. Beijing 100190, China. They are also with the University of Chinese Academy of Sciences, Beijing, China (e-mail: shaokun2014@ia.ac.cn; yuanheng.zhu@ia.ac.cn; dongbin.zhao@ia.ac.cn).}
\thanks{This work is supported by National Natural Science Foundation of China (NSFC) under Grants No.61573353, No.61603382 and No. 61533017.}
}

\maketitle

\begin{abstract}

Real-time strategy games have been an important field of game artificial intelligence in recent years.
This paper presents a reinforcement learning and curriculum transfer learning method to control multiple units in StarCraft micromanagement.
We define an efficient state representation, which breaks down the complexity caused by the large state space in the game environment.
Then a parameter sharing multi-agent gradient-descent Sarsa($\lambda$) (PS-MAGDS) algorithm is proposed to train the units.
The learning policy is shared among our units to encourage cooperative behaviors.
We use a neural network as a function approximator to estimate the action-value function, and propose a reward function to help units balance their move and attack.
In addition, a transfer learning method is used to extend our model to more difficult scenarios, which accelerates the training process and improves the learning performance.
In small scale scenarios, our units successfully learn to combat and defeat the built-in AI with 100\% win rates.
In large scale scenarios, curriculum transfer learning method is used to progressively train a group of units, and shows superior performance over some baseline methods in target scenarios.
With reinforcement learning and curriculum transfer learning, our units are able to learn appropriate strategies in StarCraft micromanagement scenarios.

\end{abstract}


\IEEEpeerreviewmaketitle

\section{Introduction}

\IEEEPARstart{A}{rtificial} intelligence (AI) has a great advance in the last decade.
As an excellent testbed for AI research, games have been helping AI to grow since its birth, including the ancient board game \cite{Silver2016Mastering},\cite{Silver2017Zero},\cite{Zhao2012Self},\cite{Shao2016Move}, the classic Atari video games\cite{Mnih2015Human},\cite{Zhao2017Deep}, and the imperfect information game\cite{Morav2017DeepStack}.
These games have a fixed, limited set of actions, and researchers only need to control a single agent in game environment.
Besides, there are a large number of games including multiple agents and requiring complex rules, which are much more difficult for AI research.

In this paper, we focus on a real-time strategy (RTS) game to explore the learning of multi-agent control.
RTS games are usually running in real-time, which is different from taking turns to play in board games\cite{Ontanon2013A}.
As one of the most popular RTS games,
StarCraft has a huge player base and numerous professional competitions, requiring different strategies, tactics and reactive control techniques.
For the game AI research, StarCraft provides an ideal environment to study the control of multiple units with different difficulty levels\cite{Lara2013A}.
In recent years, the study of StarCraft AI has an impressive progress, driven by some StarCraft AI competitions and Brood War Application Programming Interface (BWAPI)\footnote{\scriptsize http://bwapi.github.io/}\cite{Robertson2014A}.
Recently, researchers have developed more efficient platforms to promote the development of this field, including TorchCraft, ELF and PySC2.
StarCraft AI aims at solving a series of challenges, such as spatial and temporal reasoning, multi-agent collaboration, opponent modeling and adversarial planning\cite{Ontanon2013A}.
At present, designing a game AI for the full StarCraft game based on machine learning method is out-of-reach.
Many researchers focus on micromanagement as the first step to study AI in StarCraft\cite{shao2017starcraft}.
In combat scenarios, units have to navigate in highly dynamic environment and attack enemies within fire range.
There are many methods for StarCraft micromanagement, including potential fields for spatial navigation and obstacle avoidance\cite{Hagelback2016Hybrid},\cite{Uriarte2012Kiting}, bayesian modeling to deal with incompleteness and uncertainty in the game\cite{Synnaeve2016Multiscale}, heuristic game-tree search to handle both build order planning and units control\cite{Churchill2012AI}, and neuroevolution to control individual unit with hand-craft features\cite{Gabriel2012Neuroevolution}.

As an intelligent learning method, reinforcement learning (RL) is very suitable for sequential decision-making tasks.
In StarCraft micromanagement, there are some interesting applications with RL methods.
Shantia \textit{et al.} use online Sarsa and neural-fitted Sarsa with a short term memory reward function to control units' attack and retreat\cite{Shantia2011Connectionist}.
They use vision grids to obtain the terrain information. This method needs a hand-craft design, and the number of input nodes has to change with the number of units.
Besides, they apply an incremental learning method to scale the task to a larger scenario with 6 units.
However, the win rate with incremental learning is still below 50\%.
Wender \textit{et al.} use different RL algorithms in micromanagement, including $Q$ learning and Sarsa\cite{Wender2012Applying}.
They control one powerful unit to play against multiple weak units, without cooperation and teamwork between own units.

In the last few years, deep learning has achieved a remarkable performance in many complex problems\cite{Lecun2015Deep}, and has dramatically improved the generalization and scalability of traditional RL algorithms\cite{Mnih2015Human}.
Deep reinforcement learning (DRL) can teach agents to make decisions in high-dimension state space by an end-to-end method.
Usunier \textit{et al.} propose an RL method to tackle micromanagement with deep neural network\cite{Usunier2016Episodic}.
They use a greedy MDP to choose actions for units sequentially at each time step, with zero-order optimization to update the model.
This method is able to control all units owned by the player, and observe the full state of the game.
Peng \textit{et al.} use an actor-critic method and recurrent neural networks (RNNs) to play StarCraft combat games\cite{Peng2017Multiagent}.
The dependency of units is modeled by bi-directional RNNs in hidden layers, and its gradient update is efficiently propagated through the entire networks.
Different from Usunier's and Peng's work that design centralized controllers,
Foerster \textit{et al.} propose a multi-agent actor-critic method to tackle decentralized micromanagement tasks, which significantly improves the performance over centralized RL controllers\cite{Foerster2017Counterfactual}.

For StarCraft micromanagement, traditional methods have difficulties in handling complicated state and action space, and learning cooperative tactics. Modern methods rely on strong compute capability introduced by deep learning.
Besides, learning micromanagement with model-free RL methods usually needs a lot of training time, which is even more serious in large scale scenarios.
In this paper, we dedicate to explore more efficient state representation to break down the complexity caused by the large state space,
and propose appropriate RL algorithm to solve the problem of multi-agent decision making in StarCraft micromanagement.
In addition, we introduce curriculum transfer learning to extend the RL model to various scenarios and improve the sample efficiency.

The main contributions emphasize in three parts.
First, we propose an efficient state representation method to deal with the large state space in StarCraft micromanagement. This method takes units' attributes and distances into consideration, allowing an arbitrary number of units on both sides.
Compared with related work, our state representation is more concise and more efficient.
Second, we present a parameter sharing multi-agent gradient-descent Sarsa($\lambda$) (PS-MAGDS) algorithm to train our units.
Using a neural network as a function approximator, agents share the parameters of a centralized policy, and update the policy with their own experiences simultaneously.
This method trains homogeneous agents efficiently, and encourages cooperative behaviors.
To solve the problem of sparse and delayed rewards, we introduce a reward function including small intermediate rewards in the RL model. This reward function improves the training process, and serves as an intrinsic motivation to help units collaborate with each other.
Third, we propose a transfer learning method to extend our model to various scenarios. Compared with learning from scratch, this method accelerates the training process and improves the learning performance to a great extent. In large scale scenarios, we apply curriculum transfer learning method to successfully train a group of units. In term of win rates, our proposed method is superior to some baseline methods in target scenarios.

The rest of the paper is organized as follows.
In Section II, we describe the problem formulation of StarCraft micromanagement, as well as backgrounds of reinforcement learning and transfer curriculum learning.
In Section III, we present the reinforcement learning model for micromanagement, including state representation method, network architecture and action definition.
And in Section IV, we introduce the parameter sharing multi-agent gradient-descent Sarsa($\lambda$) algorithm and the reward function.
In Section V, we introduce the StarCraft micromanagement scenarios used in our paper and the training details.
In Section VI, we make an analysis of experimental results and discuss the learned strategies.
In the end, we draw a conclusion of the paper and propose some future work.

\section{Problem Formulation and Backgrounds}

\subsection{Problem Formulation}
In StarCraft micromanagement, we need to control a group of units to destroy the enemies under certain terrain conditions.
The combat scenario with multiple units is approximated as a Markov game, a multi-agent extension of Markov decision processes (MDPs)\cite{Peng2017Multiagent},\cite{Foerster2017Counterfactual}, \cite{lowe2017multi}.
In a Markov game with $N$ agents, a set of states $S$ are used to describe the properties of all agents and the environment, as well as a set of actions $A_1, ..., A_N$ and observations $O_1, ..., O_N$ for each agent.

In the combat, units in each side need to cooperate with each other.
Developing a learning model for multiple units is challenging in micromanagement.
In order to maintain a flexible framework and allow an arbitrary number of units, we consider that our units have access to the state space $S$ from its own observation of the current combat by treating other units as part of the environment $S \rightarrow O_i$.
Each unit interacts in the combat environment with its own observation and action.
$S \times A_1 \times ... \times A_N \rightarrow S'$ denotes the transition from state $S$ to the successive state $S'$ with actions of all the units,
and $R_1 ... R_N$ are the generated rewards of each unit.
For the sake of multi-agent cooperation, the policy is shared among our units.
The goal of each unit is to maximize its total expected return.

\subsection{Reinforcement Learning}
\begin{figure}[!t]
\centering
\includegraphics[width=2.5 in]{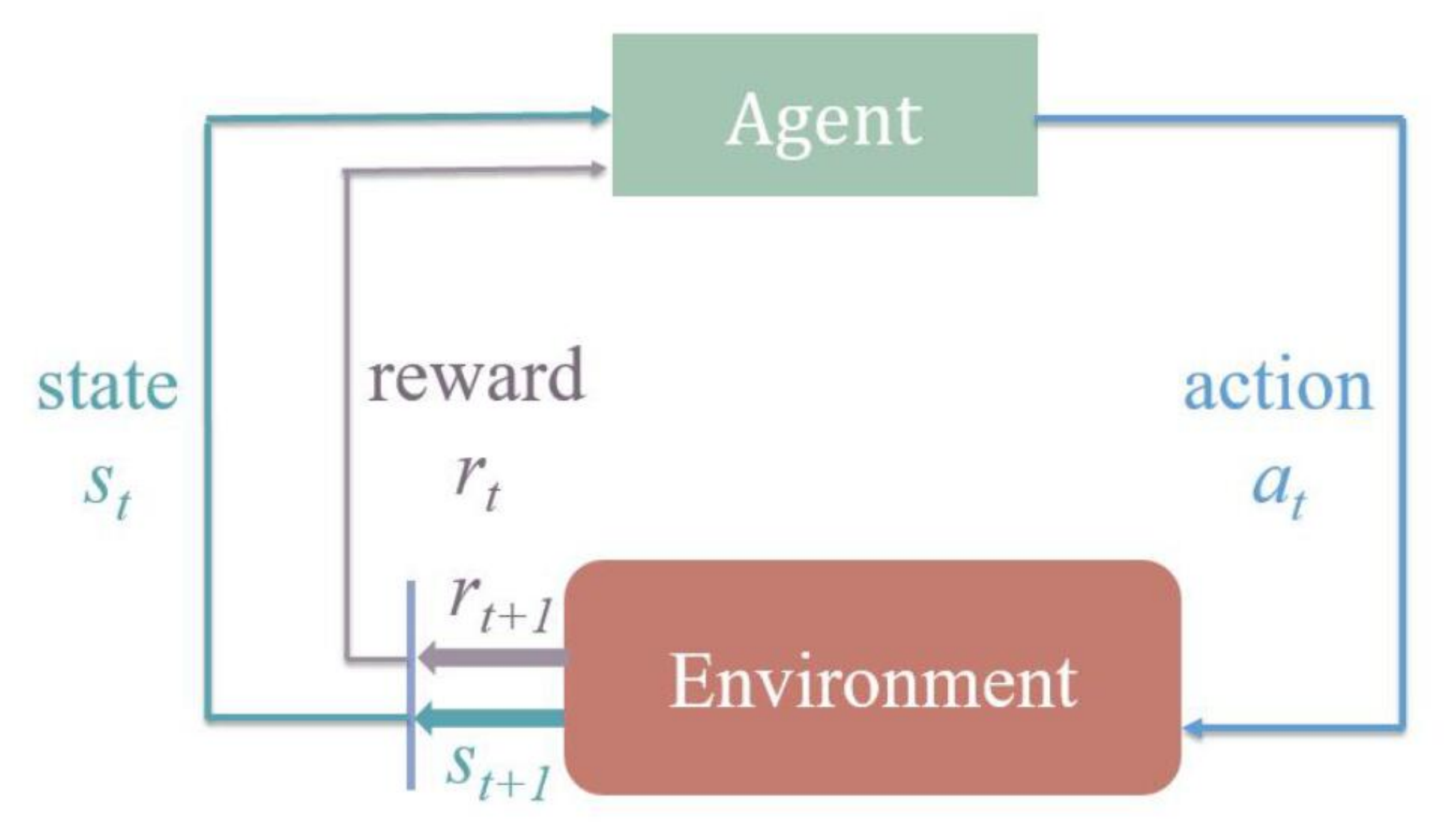}
\caption{Representation of the agent-environment interaction in reinforcement learning.}
\label{fig_sim}
\end{figure}
To solve the multi-agent control problem in StarCraft micromanagement, we can resort to reinforcement learning.
Reinforcement learning is a type of machine learning algorithms in which agents learn by trial and error and determine the ideal behavior from its own experience with the environment\cite{Sutton1998Reinforcement}.
We draw the classic RL diagram in Fig. 1. It shows the process that an RL agent interacts with the environment.
The agent-environment interaction process in RL is formulated as a Markov decision process.
The agent in state $s$ makes an action $a$ according to the policy $\pi$.
This behavior causes a reward $r$, and transfers to a new state $s'$.
We define the future discounted return at time $t$ as $\sum^T_{t'=t}\gamma^{t'-t}r_{t'}$,
where $T$ is the terminal time step and $\gamma\in[0, 1]$ is a discount factor that determines the importance of future rewards.
The aim of an RL model is to learn an optimal policy $\pi$, which defines the probability of selecting action $a$ in state $s$, so that the sum of the overall discounted rewards is maximized, as demonstrated by
\begin{eqnarray}
\max_\pi\mathbb{E}[\sum^T_{t'=t}\gamma^{t'-t}r_{t'}|s = s_t,a = a_t, \pi] .
\end{eqnarray}

As one of the most popular RL algorithms, temporal difference (TD) learning is a combination of Monte Carlo method and dynamic programming method.
TD method can learn from raw experience without a model of the environment, and update estimates based on part of the sequence, without waiting for a final outcome\cite{Kaelbling1996Reinforcement}.
The most widely known TD learning algorithms are Q-learning and Sarsa.
Q-learning estimates the value of making an action in a given state and iteratively updates the $Q$-value estimate towards the observed reward.
The TD error $\delta_t$ in Q-learning is computed as
\begin{eqnarray}
\delta_t = r_{t+1} + \gamma \max_a Q(s_{t+1},a) - Q(s_t,a_t)\ .
\end{eqnarray}

Q-learning is an off-policy learning method, which means it learns a different policy compared with the one choosing actions.
Different from Q-learning's off-policy mechanism, Sarsa is an on-policy method, which means the policy is used both for selecting actions and updating previous $Q$-value \cite{Sutton1998Reinforcement}.
The Sarsa update rule is demonstrated as
\begin{subequations}
\begin{gather}
\delta_t = r_{t+1} + \gamma Q(s_{t+1},a_{t+1}) - Q(s_t,a_t)\ , \\
Q(s_{t+1},a_{t+1}) = Q(s_t,a_t) + \alpha\delta_t\ ,
\end{gather}
\end{subequations}
where $\alpha$ is the learning rate.
Traditional reinforcement learning methods have some successful applications, including TD in Backgammon\cite{tesauro1994gammon} and adaptive dynamic programming (ADP) in control\cite{zhang2016event},\cite{zhang2017event},\cite{Zhu2017Iterative}.

Reinforcement learning with deep neural networks function approximators has received great attentions in recent years.
DRL provides an opportunity to train a single agent to solve a series of human-level tasks by an end-to-end manner\cite{Zhao2016Review}\cite{tang2018Recent}.
As the most famous DRL algorithm, deep Q-network (DQN) uses the experience replay technique and a target network to remove the correlations between samples and stabilize the training process\cite{Mnih2015Human}.
In the last few years, we have witnessed a great number of improvements on DQN, including double DQN\cite{van2016deep}, prioritised DQN\cite{schaul2016prioritized}, dueling DQN\cite{wang2016dueling}, distributed DQN\cite{nair2015massively} and asynchronous DQN\cite{mnih2016asynchronous}.
Apart from value-based DRL methods like DQN and its variants, policy-based DRL methods use deep networks to parameterize and optimize the policy directly\cite{li2017policy}.
Deep deterministic policy gradient (DDPG) is the continuous analogue of DQN, which uses a critic to estimate the value of current policy and an actor to update the policy\cite{lillicrap2016continuous}.
Policy-based DRL methods play important roles in continuous control, including asynchronous advantage actor-critic (A3C)\cite{mnih2016asynchronous}, trust region policy optimization (TRPO)\cite{schulman2015trust}, proximal policy optimization (PPO)\cite{Schulman2017Proximal}, and so on.
The sample complexity of traditional DRL methods tends to be high, which limits these methods to real-world applications.
While model-based DRL approaches learn value function and policy in a data-efficient way, and have been widely used in sensorimotor control.
Guided policy search (GPS) uses a supervised learning algorithm to train policy and an RL algorithm to generate guiding distributions, allowing to train deep policies efficiently \cite{levine2013guided}.
Researchers have also proposed some other model-based DRL methods, like normalized advantage functions (NAF)\cite{gu2016continuous} and embed to control (E2C)\cite{watter2015embed}.

Multi-agent reinforcement learning is a closely related area to our work\cite{littman1994markov}.
A multi-agent system includes a number of agents interacting in one environment\cite{ming1993multi}\cite{Zhang2017FMRQ}.
Recently, some multi-agent reinforcement learning algorithms with deep neural network are proposed to learn communication\cite{sukhbaatar2016learning}, cooperative-competitive behaviors\cite{lowe2017multi} and imperfect information\cite{marc2017unified}.
In our work, we use a multi-agent reinforcement learning method with policy sharing among agents to learn cooperative behaviors.
Agents share the parameters of a centralized policy, and update the policy with their own experiences simultaneously.
This method can train homogeneous agents more efficiently\cite{jayesh2017cooperative}.
\begin{figure}[!t]
\centering
\includegraphics[width=3 in]{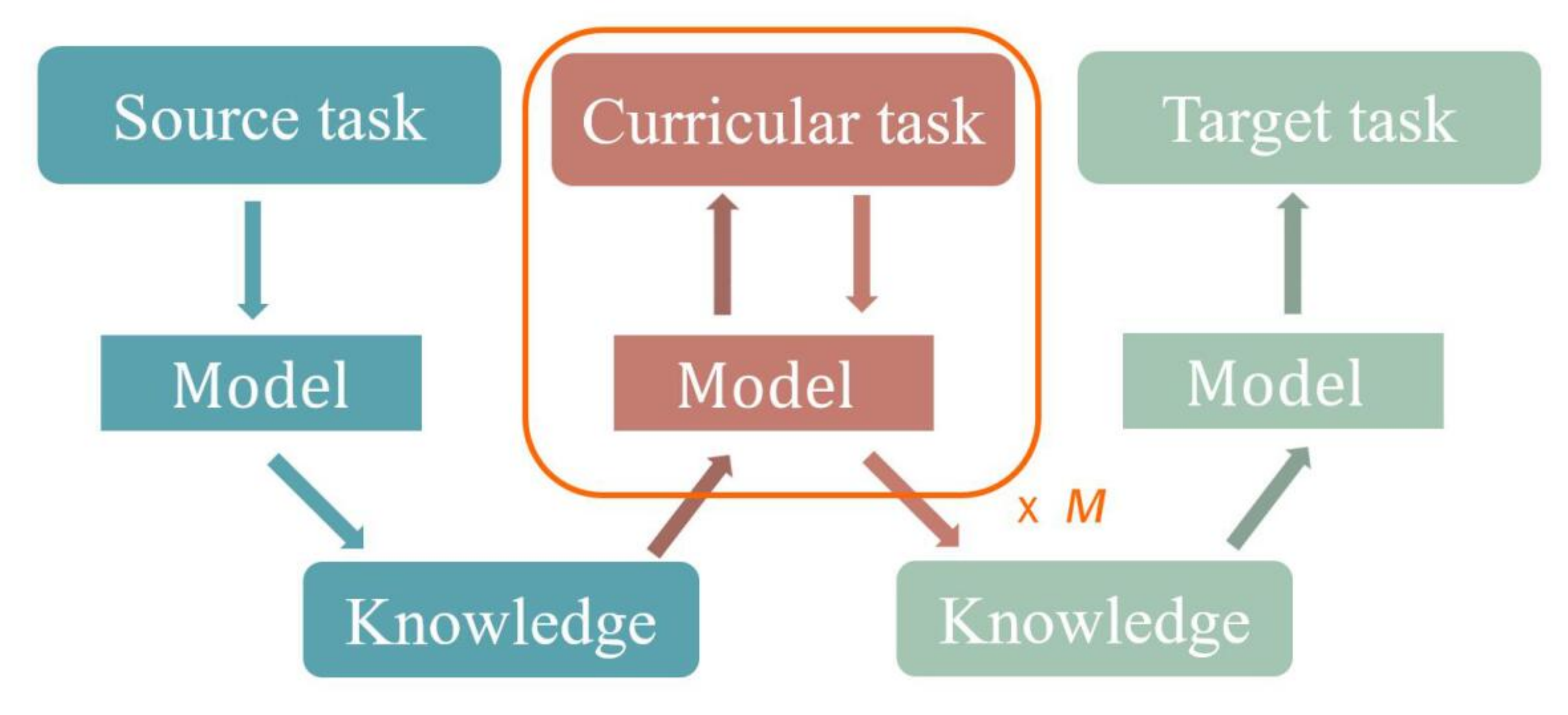}
\caption{Representation of curriculum transfer learning. Storing knowledge gained solving source task and gradually applying it to $M$ curricular tasks to update the knowledge. Eventually, applying it to the target task.}
\label{fig_sim}
\end{figure}

\subsection{Curriculum Transfer Learning}
\begin{table*}[!t]
\renewcommand{\arraystretch}{1.3}
\caption{The data type and dimension of inputs in our model.}
\label{table_example}
\centering
\begin{tabular}{c c c c c c c c c}
\hline
\hline
Inputs & CoolDown & HitPoint &  OwnSumInfo & OwnMaxInfo & EnemySumInfo & EnemyMaxInfo & TerrainInfo & Action \\
\hline
Type & $\in R$ & $\in R$  & $\in R$  & $\in R$ & $\in R$ & $\in R$ & $\in R$ & $\in cat. $ \\
Dimension & 1 & 1 & 8 & 8 & 8 & 8 & 8 & 9 \\
\hline
\hline
\end{tabular}\\
\ \\
\small Note: $ R $  means the input has real value and $ cat.$  means the input is categorical and one-hot encoded.
\end{table*}

Generally speaking, model-free reinforcement learning methods need plenty of samples to learn an optimal policy.
However, many challenging tasks are difficult for traditional RL methods to learn admissible policies in large state and action space.
In StarCraft micromanagement, there are numerous scenarios with different units and terrain conditions.
It will take a lot of time to learn useful strategies in different scenarios from scratch.
A number of researchers focus on improving the learning speed and performance by exploiting domain knowledge across various but related tasks.
The most widely used approach is transfer learning (TL)\cite{Pan2010A}\cite{Gupta2018transfer}.
To some extent, transfer learning is a kind of generalization across tasks, transferring knowledge from source tasks to target tasks.
Besides, transfer learning can be extended to RL problems by using the model parameters in the same model architecture\cite{Taylor2009Transfer}.
The procedure of using transfer learning in our experiments is training the model with RL method in a source scenario first.
And then, we can use the well-trained model as a starting point to learn micromanagement in target scenarios.

As a special form of transfer learning, curriculum learning involves a set of tasks organized by increasing level of difficulties.
The initial tasks are used to guide the learner so that it can perform better on the final task\cite{Bengio2009Curriculum}.
Combining curriculum learning and transfer learning, curriculum transfer learning (CTL) method has shown good performance to help the learning process converge faster and towards better optimum in recent work\cite{Graves2016Hybrid},\cite{Wu2017Training},\cite{dong2017multi-task}.
For micromanagement, a feasible method of using CTL is mastering a simple scenario first, and then solving difficult scenarios based on this knowledge.
By changing the number and type of units, we could control the difficulty of micromanagement.
In this way, we can use CTL to train our units with a sequence of progressively difficult micromanagement scenarios, as shown in Fig. 2.

\section{Learning Model for Micromanagement}
\subsection{Representation of High-Dimension State}

State representation of StarCraft is still an open problem with no universal solution.
We construct a state representation with inputs from the game engine, which have different data types and dimensions, as depicted in Table I.
The proposed state representation method is efficient and independent of the number of units in the combat.
In summary, the state representation is composed of three parts: the current step state information, the last step state information and the last step action, as shown in Fig. 3.
The current step state information includes own weapon's cooldown time, own unit's hitpoint, distances information of own units,
distances information of enemy units and distances information of terrain.
The last step state information is the same with the current step.
We take the last step action into consideration, which has been proven to be helpful for the learning process in the RL domain\cite{Wang2017Learning},\cite{Mirowski2017Learning}.
The proposed state representation method also has good generalization and can be used in other combat games, which need to take agents' property and distance information into consideration.

All inputs with real type are normalized by their maximum values. Among them, CoolDown and HitPoint have 1 dimension for each.
We divide the combat map into 8 sector areas on average, and compute the distances information in each area.
Units' distance information is listed as follows:
\begin{itemize}
\item OwnSumInfo: own units' distances are summed in each area;
\item OwnMaxInfo: own units' distances are maximized in each area;
\item EnemySumInfo: enemy units' distances are summed in each area;
\item EnemyMaxInfo: enemy units' distances are maximized in each area.
\end{itemize}

If a unit is out of the central unit's sight range $D$, the unit's distance value $dis\_unit$ is set to 0.05. Otherwise, the value is linear with $d$, the distance to the central unit, as demonstrated in equation (4).
\begin{equation}
dis\_unit(d)=\begin{cases}
0.05,\quad d > D \\
1 - 0.95(d/D),\  d\leq D
\end{cases}
\end{equation}

In addition, terrain distance value $dis\_terrain$ is also computed in 8 sector areas.
If the obstacle is out of the central unit's sight range, the value is set to 0. Otherwise, the value is also linear with the distance to the central unit, as shown in equation (5).
\begin{equation}
dis\_terrain(d)=\begin{cases}
0,\quad d > D \\
1 - d/D, \quad d \leq D
\end{cases}
\end{equation}

In this way, the current step state information has 42 dimensions.
The last step action has 9 dimensions, with selected action setting to 1 and the other actions setting to 0.
In total, the state representation in our model is embedded to 93 dimensions.
\begin{figure}[!t]
\centering
\includegraphics[width=2.9 in]{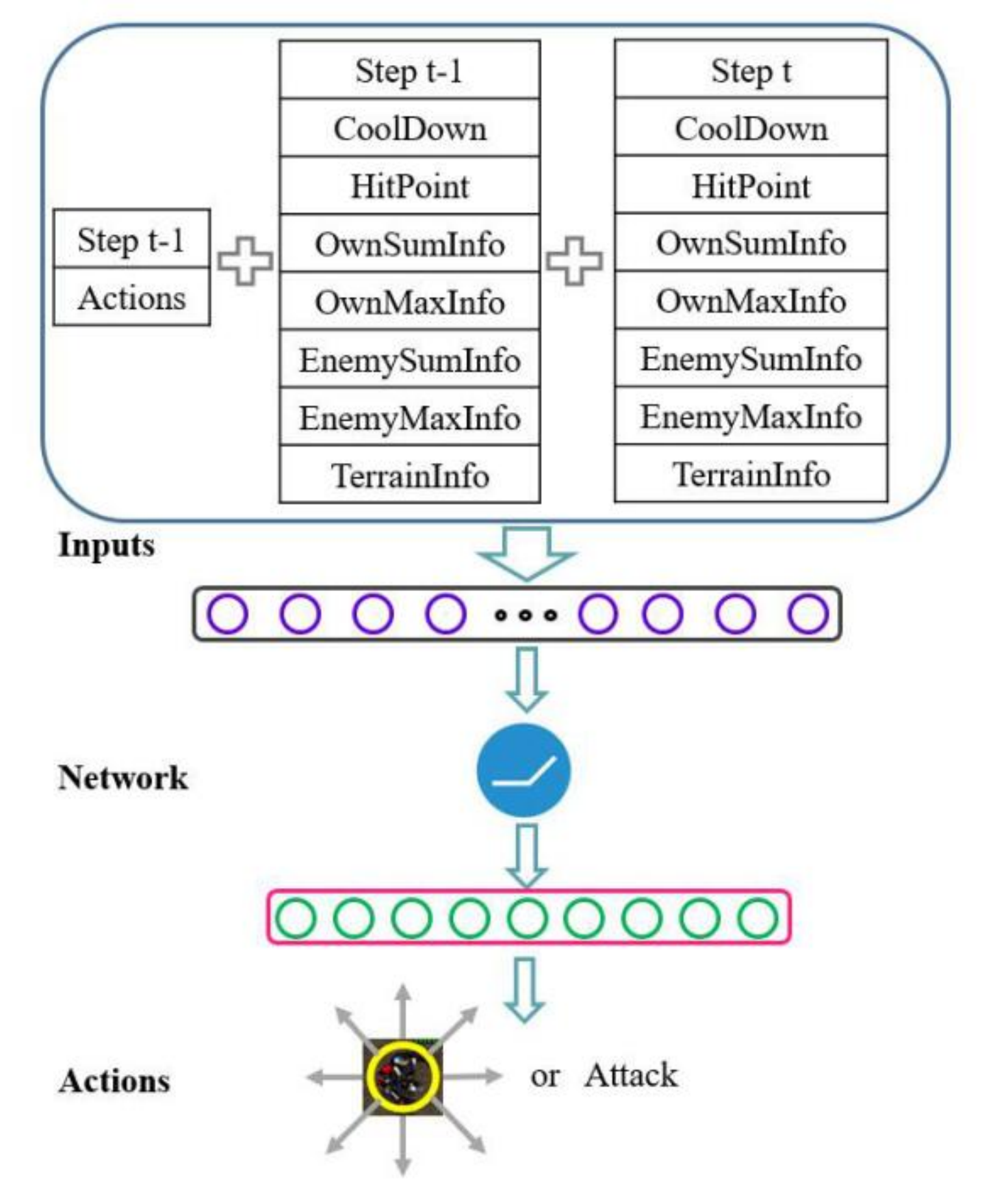}
\caption{Representation of the learning model of one unit in StarCraft micromanagement scenarios. The state representation has three parts and a neural network is used as the function approximator. The network outputs the probabilities of moving to 8 directions and attack.}
\label{fig_sim}
\end{figure}

\subsection{Action Definition}
In StarCraft micromanagement scenarios, the original action space is very large.
At each time step, each unit can move to arbitrary directions with arbitrary distances in the map.
When the unit decides to attack, it can choose any enemy units in the weapon's fire range.
In order to simplify the action space, we choose 8 move directions with a fixed distance and attacking the weakest as the available actions for each unit.
When the chosen action is move, our units will turn to one of the 8 directions, Up, Down, Left, Right, Upper-left, Upper-right, Lower-left, Lower-right, and move a fixed distance.
When the chosen action is attack, our units will stay at the current position and focus fire on enemy units.
Currently, we select the enemy with the lowest hitpoint in our weapon's attack range as the target.
According to the experimental results, these actions are enough to control our units in the game.

\subsection{Network Architecture }
Because our units' experience has a limited subset of the large state space and most test states will never been explored before,
it will be difficult to apply table reinforcement learning to learn an optimal policy.
To solve this problem, we use a neural network parameterized by vector $\boldsymbol{\theta}$
to approximate the state-action values to improve our RL model's generalization.

The input of the network is the 93 dimensions tensor from the state representation.
We has 100 neurons in the hidden layer, and use the rectified linear unit (ReLU) activation function for the network nonlinearity, as demonstrated
by
\begin{eqnarray}
f(\boldsymbol{z}) = \max(\boldsymbol{0}, \boldsymbol{z}),
\end{eqnarray}
where $\boldsymbol{z}$ is the output of hidden layer.
We use ReLU function rather than Sigmoid or tanh function, because ReLU function does not have the problem of gradient descent, which can guarantee the effective training of the model\cite{Glorot2011Deep}.
Different from these saturating nonlinearities functions such as Sigmoid or tanh, ReLU function is a non-saturating nonlinearity function.
In terms of training time with gradient descent, the non-saturating nonlinearity is much faster\cite{Nair2010Rectified}.

The output layer of the neural network has 9 neurons, giving the probabilities of moving to 8 directions and attack.
The learning model of one unit in StarCraft micromanagement scenarios, including state representation, neural network architecture and output actions, is depicted in Fig. 3.

\section{Learning Method for Micromanagement}

In this paper, we formulate StarCraft micromanagement as a multi-agent reinforcement learning model.
We propose a parameter sharing multi-agent gradient-descent Sarsa($\lambda$) (PS-MAGDS) method to train the model,
and design a reward function as intrinsic motivations to promote the learning process.
The whole PS-MAGDS reinforcement learning diagram is depicted in Fig. 4.

\subsection{Parameter Sharing Multi-agent Gradient-Descent Sarsa($\lambda$)}
\begin{figure}[!t]
\centering
\includegraphics[width=3.4 in]{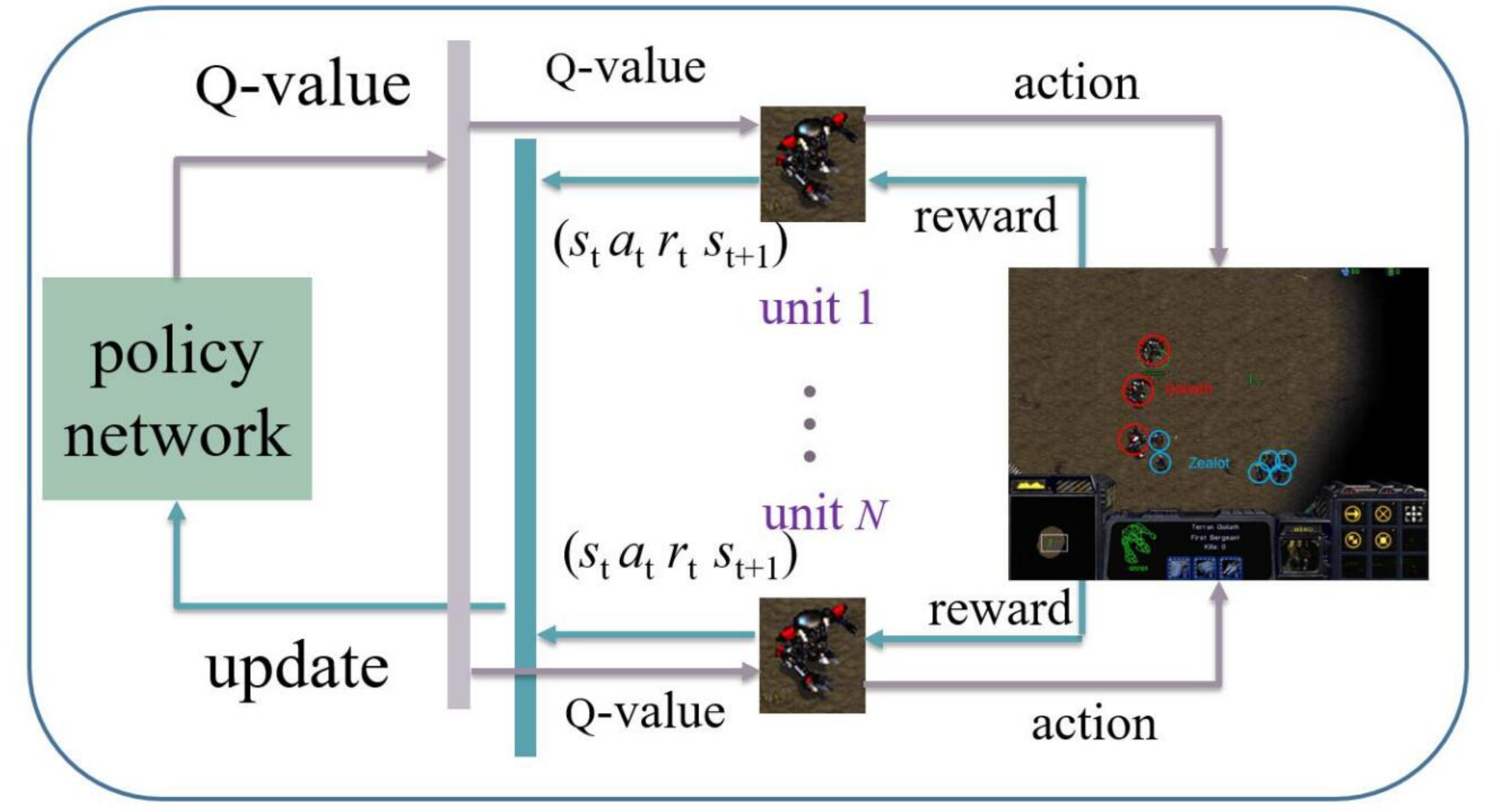}
\caption{The PS-MAGDS reinforcement learning diagram in the StarCraft micromanagement scenarios.}
\label{fig_sim}
\end{figure}

We propose a multi-agent RL algorithm that extend the traditional Sarsa($\lambda$) to multiple units by sharing the parameters of the policy network among our units.
To accelerate the learning process and tackle the problem of delayed rewards, we use eligibility traces in reinforcement learning.
As a basic mechanism in RL, eligibility traces are used to assign temporal credit,
which consider a set of previously experienced transitions\cite{Singh1996Reinforcement}.
This means it not only considers the value of the last state-action pair but also the visited ones.
With this method, we can solve the problem of delayed reward in the game environment.
Sarsa with eligibility traces, termed as Sarsa($\lambda$), is one way of averaging backups made after multiple steps.
$\lambda$ is a factor that determines the weight of each backup.
In our implementation of Sarsa($\lambda$) for multiple units combat, we use a neural network as the function approximator and share network parameters among all our units.
Although we have only one network to train,
the units can still behave differently because each one receives different observations and actions as its input.

To update the policy network efficiently, we use the gradient-descent method to train the Sarsa($\lambda$) reinforcement learning model.
The gradient-descent learning update is demonstrated in equation (7),
\begin{subequations}
\begin{gather}
\delta_t = r_{t+1}+\gamma Q(s_{t+1}, a_{t+1};\boldsymbol{\theta}_{t}) - Q(s_t, a_t;\boldsymbol{\theta}_t)\\
\boldsymbol{\theta}_{t+1} = \boldsymbol{\theta}_t + \alpha\delta_t \boldsymbol{e}_t\\
\boldsymbol{e}_t=\gamma\lambda \boldsymbol{e}_{t-1}+\nabla_{\boldsymbol{\theta}_t} Q(s_t, a_t;\boldsymbol{\theta}_t), \boldsymbol{e}_0=\boldsymbol{0}
\end{gather}
\end{subequations}
where $\boldsymbol{e}_t$ is the eligibility traces at time step $t$.

One of the challenging issues in reinforcement learning is the trade-off between exploration and exploitation.
If we choose the best action every step according to current policy, we are likely to trap in local optimum.
On the contrary, if we tend to explore in the large state space, the model will have difficulty in converging.
In the experiment, we use the $\epsilon$-greedy method to choose actions during training, which selects the current best action with probability $1-\epsilon$, and takes a random exploratory action with probability $\epsilon$,
\begin{equation}
a=\begin{cases}
randint(N),\quad random(0,1)< \epsilon \\
\arg \mathop{\max}_a Q(s, a),\quad otherwise
\end{cases}
\end{equation}
where $N$ equals to 9 in the experiment.

We use exponentially $\epsilon$ decay to implement the $\epsilon$-greedy method.
The $\epsilon$ is initialized to 0.5 and anneals schedule with an exponential smoothing window of the episode number $episode\_num$, as demonstrated by
\begin{eqnarray}
\epsilon = 0.5/\sqrt{1+episode\_num}\ .
\end{eqnarray}

The overall parameter sharing multi-agent gradient-descent Sarsa($\lambda$) method is presented in Algorithm 1.
\subsection{Reward Function}

The reward function provides useful feedbacks for RL agents, which has a significant impact on the learning results\cite{Ng1999Policy}.
The goal of StarCraft micromanagement is to destroy all of the enemy units in the combat.
If the reward is only based on the final result, the reward function will be extremely sparse.
Moreover, units usually get a positive reward after many steps.
The delay in rewards makes it difficult to learn which set of actions is responsible for the corresponding rewards.

To tackle the problem of sparse and delayed rewards in micromanagement, we design a reward function to include small intermediate rewards.
In our experiment, all agents receive the main reward caused by their attack action at each time step, equalling to the damage that the enemy units received minus the hitpoint loss of our units.
\begin{eqnarray}
\begin{split}
r_t = (damage\_amount_t \times damage\_factor - \rho \times \\
(unit\_hitpoint_{t-1} - unit\_hitpoint_t))/10
\end{split}
\end{eqnarray}
where $damage\_amount$ is the amount of damage caused by our units' attack, $damage\_factor$ is our units' attack force and $unit\_hitpoint$ is our unit's hitpoint.
We divide the reward by a constant to resize it to a more suitable range, which is set to 10 in our experiment.
$\rho$ is a normalized factor to balance the total hitpoint of our units and enemy units,
\begin{eqnarray}
\begin{split}
\rho = \sum_{i=1}^{H} enemy\_hitpoint_i / \sum_{j=1}^{N} unit\_hitpoint_j
\end{split}
\end{eqnarray}
where $H$ is the number of enemy units, and $N$ is the number of our units.
Generally speaking, this normalized factor is necessary in StarCraft micromanagement with different numbers and types of units.
Without proper normalization, policy network will have difficulty in converging,
and our units need much more episodes to learn useful behaviors.

Apart from the basic attack reward, we consider some extra rewards as the intrinsic motivation to speed up the training process.
When a unit is destroyed, we introduce an extra negative reward, and set it to -10 in our experiment.
We would like to punish this behavior in consideration that the decrease of own units has a bad influence on the combat result.
Besides, in order to encourage our units to work as a team and make cooperative actions, we introduce a reward for units' move.
If there are no our units or enemy units in the move direction, we give this move action a small negative reward, which is set to -0.5.
According to the experiment, this reward has an impressive effect on the learning performance, as shown in Fig. 6.

\begin{algorithm}[!t]
\caption{Parameter Sharing Multi-Agent Gradient-Descent Sarsa($\lambda$)}
\begin{algorithmic}[1]
\STATE Initialize policy parameters $\boldsymbol{\theta}$ shared among our units
\STATE Repeat (for each episode):
\STATE \ \ \ $\boldsymbol{e}_0=\boldsymbol{0}$
\STATE \ \ \ Initialize $s_t$, $a_t$
\STATE \ \ \ Repeat (for each step of episode):
\STATE \ \ \ \ \ \ Repeat (for each unit):
\STATE \ \ \ \ \ \ \ \ \ Take action $a_t$, receive $r_{t+1}$, next state $s_{t+1}$
\STATE \ \ \ \ \ \ \ \ \ Choose $a_{t+1}$ from $s_{t+1}$ using $\epsilon$-greedy
\STATE \ \ \ \ \ \ \ \ \ If $random(0,1)< \epsilon$
\STATE \ \ \ \ \ \ \ \ \ \ \ \ $a_{t+1} = randint(N)$
\STATE \ \ \ \ \ \ \ \ \ else
\STATE \ \ \ \ \ \ \ \ \ \ \ \ $a_{t+1} = \arg\mathop{\max}_{a} Q(s_{t+1}, a;\boldsymbol{\theta}_{t})$
\STATE \ \ \ \ \ \ Repeat (for each unit):
\STATE \ \ \ \ \ \ \ \ \ Update TD error, weights and eligibility traces
\STATE \ \ \ \ \ \ \ \ \ $\delta_t = r_{t+1}+\gamma Q(s_{t+1}, a_{t+1};\boldsymbol{\theta}_{t}) - Q(s_t, a_t;\boldsymbol{\theta}_t)$
\STATE \ \ \ \ \ \ \ \ \ $\boldsymbol{\theta}_{t+1} = \boldsymbol{\theta}_t + \alpha\delta_t \boldsymbol{e}_t$
\STATE \ \ \ \ \ \ \ \ \ $\boldsymbol{e}_{t+1}=\gamma\lambda \boldsymbol{e}_t+\nabla_{\boldsymbol{\theta}_{t+1}} Q(s_{t+1}, a_{t+1};\boldsymbol{\theta}_{t+1})$
\STATE \ \ \ \ \ \ \ \ \ $t \leftarrow t+1$
\STATE \ \ \ until $s_t$ is terminal
\end{algorithmic}
\end{algorithm}

\subsection{Frame Skip}

When applying reinforcement learning to video games, we should pay attention to the continuity of actions.
Because of the real-time property of StarCraft micromanagement, it is impractical to make a action every game frame.
One feasible method is using frame skip technology, which executes a training step every fixed number of frames.
However, small frame skip will introduce strong correlation in the training data, while large frame skip will reduce effective training samples.
We refer to related work in \cite{Usunier2016Episodic}, and try several frame skips (8, 10, 12) in a small scale micromanagement scenario.
At last, we set the frame skip to 10 in our experiment, which takes an action every 10 frames for each unit.

\section{Experiment settings}
\begin{figure*}[htbp]
\centering
\subfigure{
\begin{minipage}{5.5cm}
\centering
\includegraphics[width=2.1 in, trim=50 0 0 0]{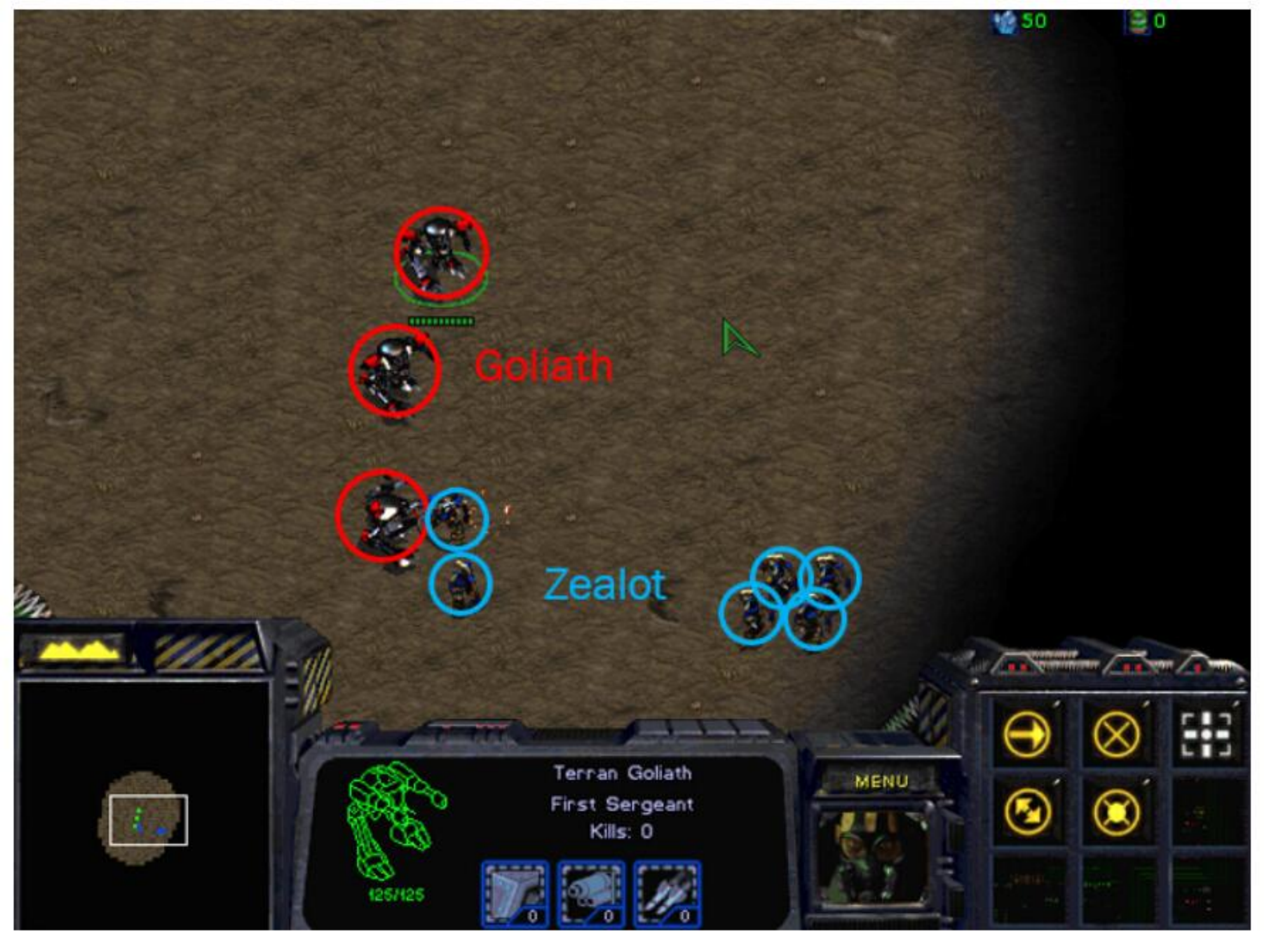}
\end{minipage}
}
\subfigure{
\begin{minipage}{5.5 cm}
\centering
\includegraphics[width=2.15 in, trim=50 0 0 0]{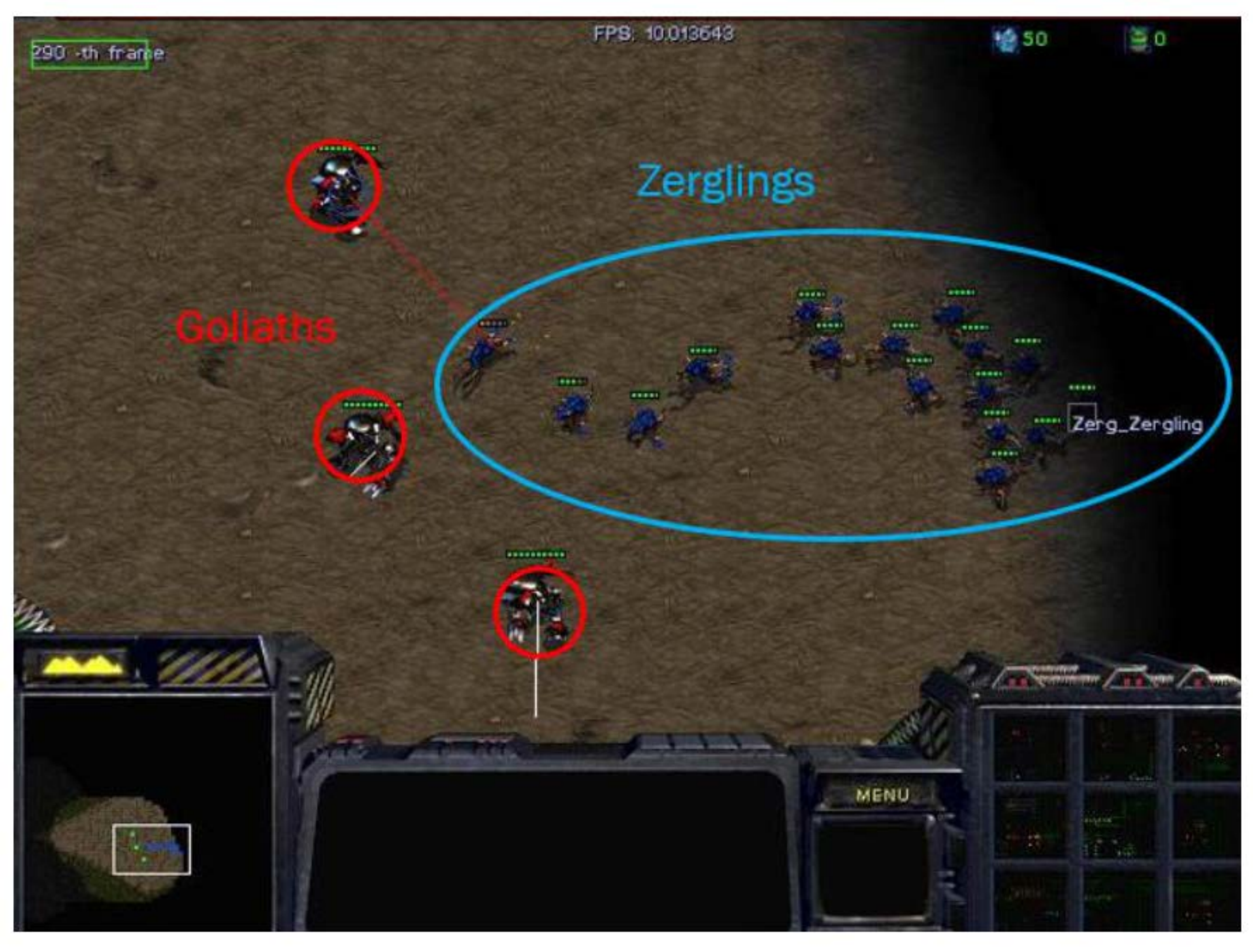}
\end{minipage}
}
\subfigure{
\begin{minipage}{5.5cm}
\centering
\includegraphics[width=2.38 in, trim=47 0 0 0]{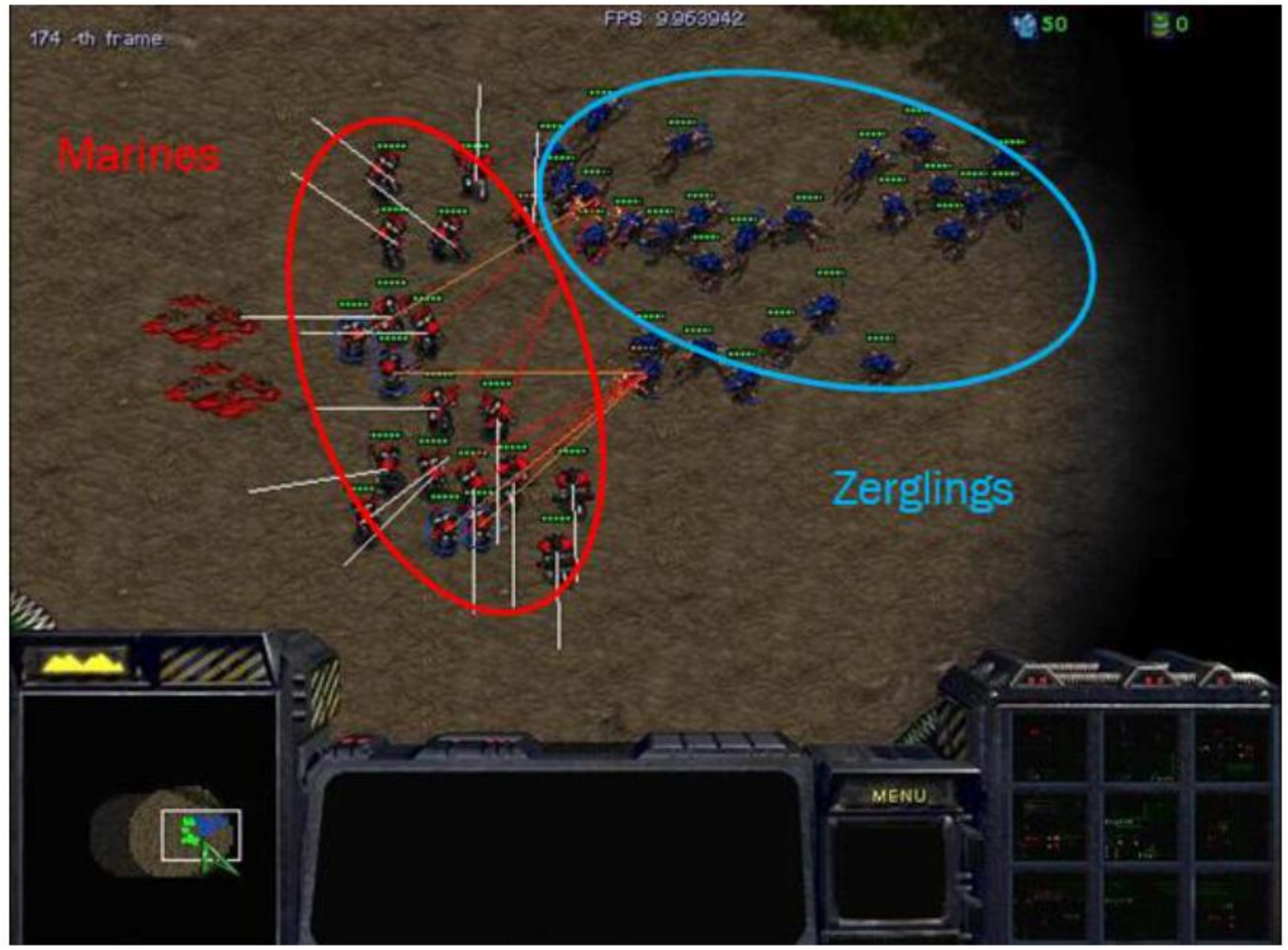}
\end{minipage}
}
\caption{Representation of the StarCraft micromanagement scenarios in our experiments, left: Goliaths vs. Zealots; middle: Goliaths vs. Zerglings; right: Marines vs. Zerglings.}
\label{fig_sim}
\end{figure*}

\begin{table}[!t]
\renewcommand{\arraystretch}{1.3}
\caption{The comparative attributes of different units in our micromanagement scenarios.}
\label{table_example}
\centering
\begin{tabular}{c c c c c}
\hline
\hline
Attributes & Goliath & Zealot & Zergling & Marine \\
\hline
Race & Terran & Protoss & Zerg & Terran \\
HitPoint & 125 & 160 & 35 & 40 \\
CoolDown & 22 & 22 & 8 & 15 \\
Damage Factor & 12 & 16 & 5 & 6 \\
Defence Factor & 1 & 1 & 0 & 0\\
Fire Range & 5 & 1 & 1 & 4 \\
Sight Range & 8 & 7 & 5 & 7 \\
\hline
\hline
\end{tabular}
\end{table}

\subsection{StarCraft Micromanagement Scenarios}

We consider several StarCraft micromanagement scenarios with various units, including Goliaths vs. Zealots, Goliaths vs. Zerglings and Marines vs. Zerglings,
as shown in Fig. 5.

\begin{itemize}
\item[1)] In the first scenario, we will control 3 Goliaths to fight against 6 Zealots. From Table II, we can see that the enemy units have advantage on the number of units, hitpoint and damage factor. By contrast, our units' fire range is much wider.
\item[2)] In the second scenario, the enemies have 20 Zerglings. Our Goliaths units have advantage on hitpoint, damage factor and fire range, while the enemies have much more units and less cooldown time.
\item[3)] In the third scenario, we will control up to 20 Marines to fight against 30 Zerglings. The enemy units have advantage on speed and amount, while our units have advantage on fire range and damage factor.
\end{itemize}

We divide these scenarios into two groups. The first and the second are small scale micromanagements and the last is the large scale micromanagement.
In these scenarios, the enemy units are controlled by the built-in AI, which is hard-coded with game inputs.
An episode terminates when either side of the units are destroyed.
A human beginner of StarCraft can't beat the built-in AI in these scenarios. Platinum-level players have average win rates of below 50\% with 100 games for each scenario.
Our RL agents is expected to exploit their advantages and avoid their disadvantages to win these combats.

\subsection{Training}

In the training process,
we set the discount factor $\gamma$ to 0.9, the learning rate $\alpha$ to 0.001, and the eligibility traces factor $\lambda$ to 0.8 in all scenarios.
Moreover, the maximum steps of each episode are set to 1000.
In order to accelerate the learning process, the game runs at full speed by setting gameSpeed to 0 in BWAPI.
The experiment is applied on a computer with an Intel i7-6700 CPU and 16GB of memory.

\section{Results And Discussions}

In this section, we analyze the results in different micromanagement scenarios and discuss our RL model's performance.
In small scale scenarios, we use the first scenario as a starting point to train our units.
In the remaining scenarios, we introduce transfer learning method to scale the combat to large scenarios.
The object of StarCraft micromanagement is defeating the enemies and increasing the win rates in these given scenarios.
For a better comprehension, we analyze the win rates, episode steps and average rewards during training, as well as the learned strategies.
Our code and results are open-source at https://github.com/nanxintin/StarCraft-AI.

\begin{figure}[!t]
\centering
\includegraphics[width=3.3 in, trim=0 0 0 0]{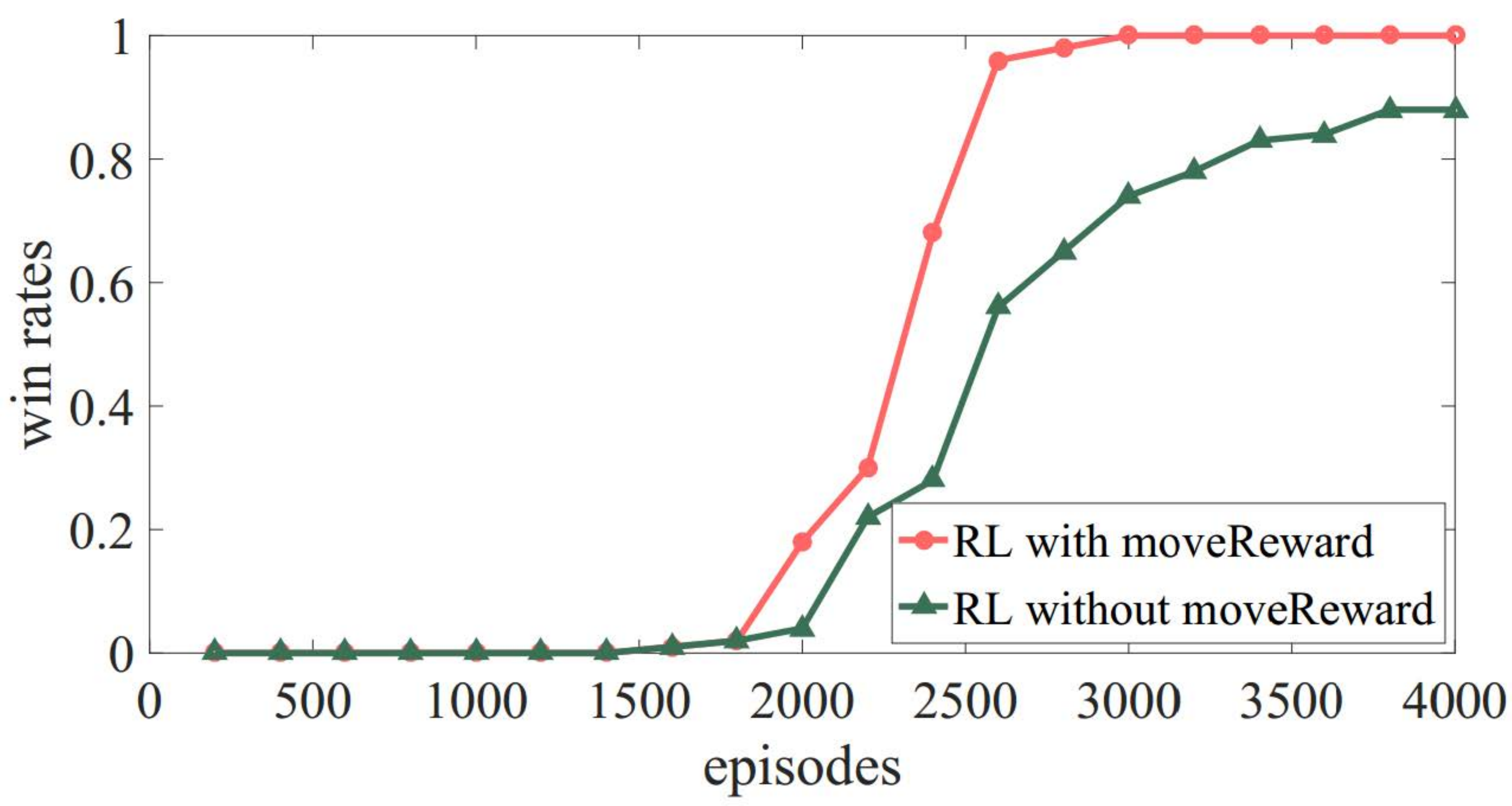}
\caption{The win rates of our units in 3 Goliaths vs. 6 Zealots micromanagement scenario from every 200 episodes' training.}
\label{fig_sim}
\end{figure}

\subsection{Small Scale Micromanagement}
In small scale micromanagement scenarios, we will train Goliaths against enemy units with different amounts and types.
In the second scenario, we will also use transfer learning method to train Goliaths based on the well-trained model of the first scenario.
Both of the two scenarios are trained with 4000 episodes and over 1 million steps.

\subsubsection{Goliaths vs. Zealots}

In this scenario, we train our Goliaths units from scratch and analyze the results.
\begin{itemize}
\item Win Rates: At first, we will analyze the learning performance of our RL method with moveReward. To evaluate the win rates, we test our model after every 200 episodes' training for 100 combats, and depict the results in Fig. 6. We can see that our Goliaths units can't win any combats before 1400 episodes. With the progress of training, units start to win several games and the curve of win rates has an impressive increase after 2000 episodes. After 3000 episodes' training, our units can reach win rates of $100\%$ at last.
\item Episode Steps: We depict the average episode steps and standard deviations of our three Goliaths units during training in Fig. 7. It is apparent to see that the curve of average episode steps has four stages. In the opening, episode steps are extremely few because Goliaths have learned nothing and are destroyed quickly. After that, Goliaths start to realize that the hitpoint damage causes a negative reward. They learn to run away from enemies and the episode steps increase to a high level. And then, episode steps start to decrease because Goliaths learn to attack to get positive rewards, rather than just running away. In the end, Goliaths have learned an appropriate policy to balance move and attack, and they are able to destroy enemies in almost 300 steps.
\item Average Rewards: Generally speaking, a powerful game AI in micromanagement scenarios should defeat the enemies as soon as possible. Here we introduce the average rewards, dividing the total rewards by episode steps in the combat. The curve of our Goliaths units' average rewards is depicted in Fig. 8. The average rewards have an obvious increase in the opening, grow steadily during training and stay smooth after almost 3000 episodes.
\end{itemize}

\begin{figure}[!t]
\centering
\includegraphics[width=3.3 in]{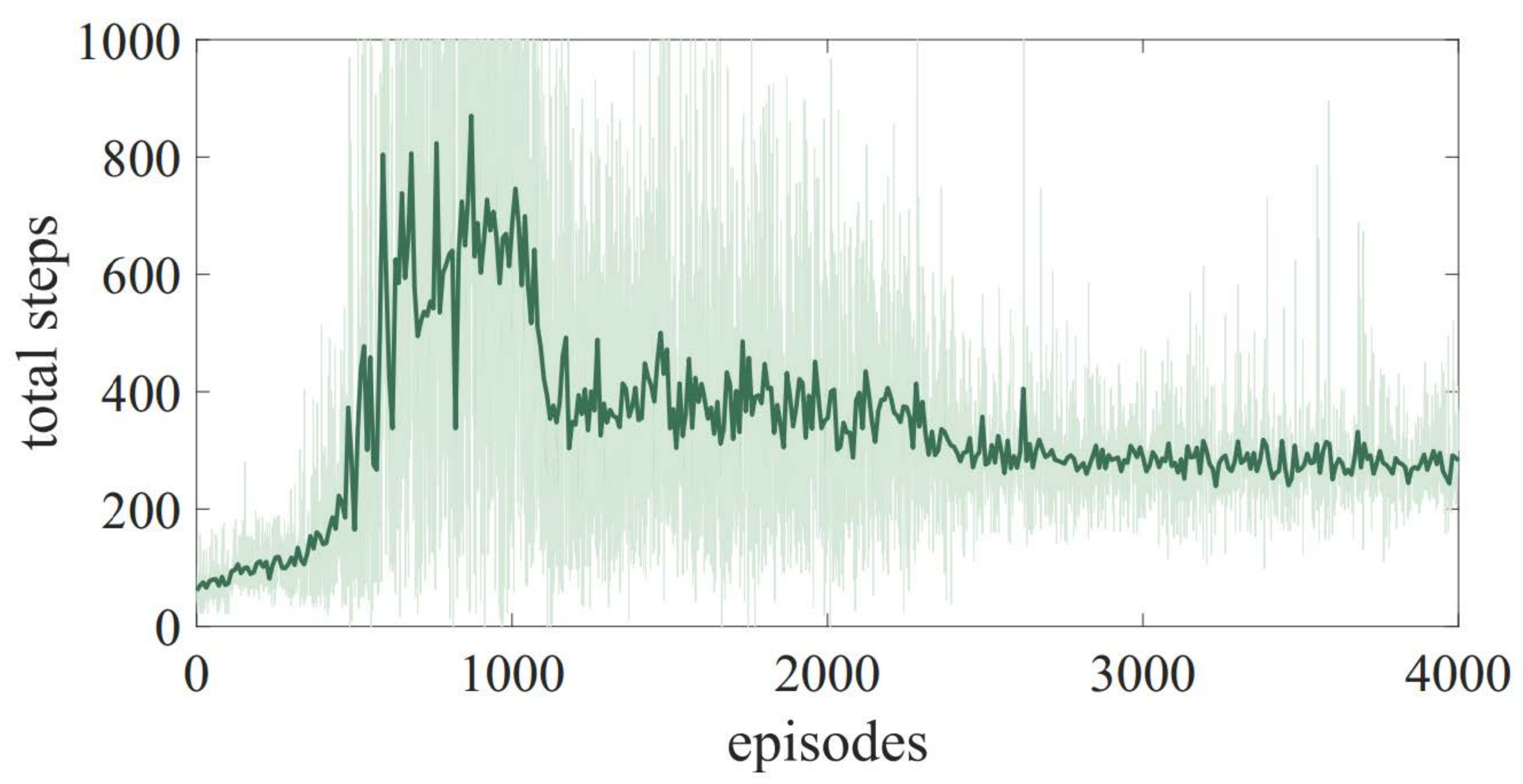}
\caption{The episode steps of our units in 3 Goliaths vs. 6 Zealots micromanagement scenario during training.}
\label{fig_sim}
\end{figure}

\begin{figure}[!t]
\centering
\includegraphics[width=3.3 in]{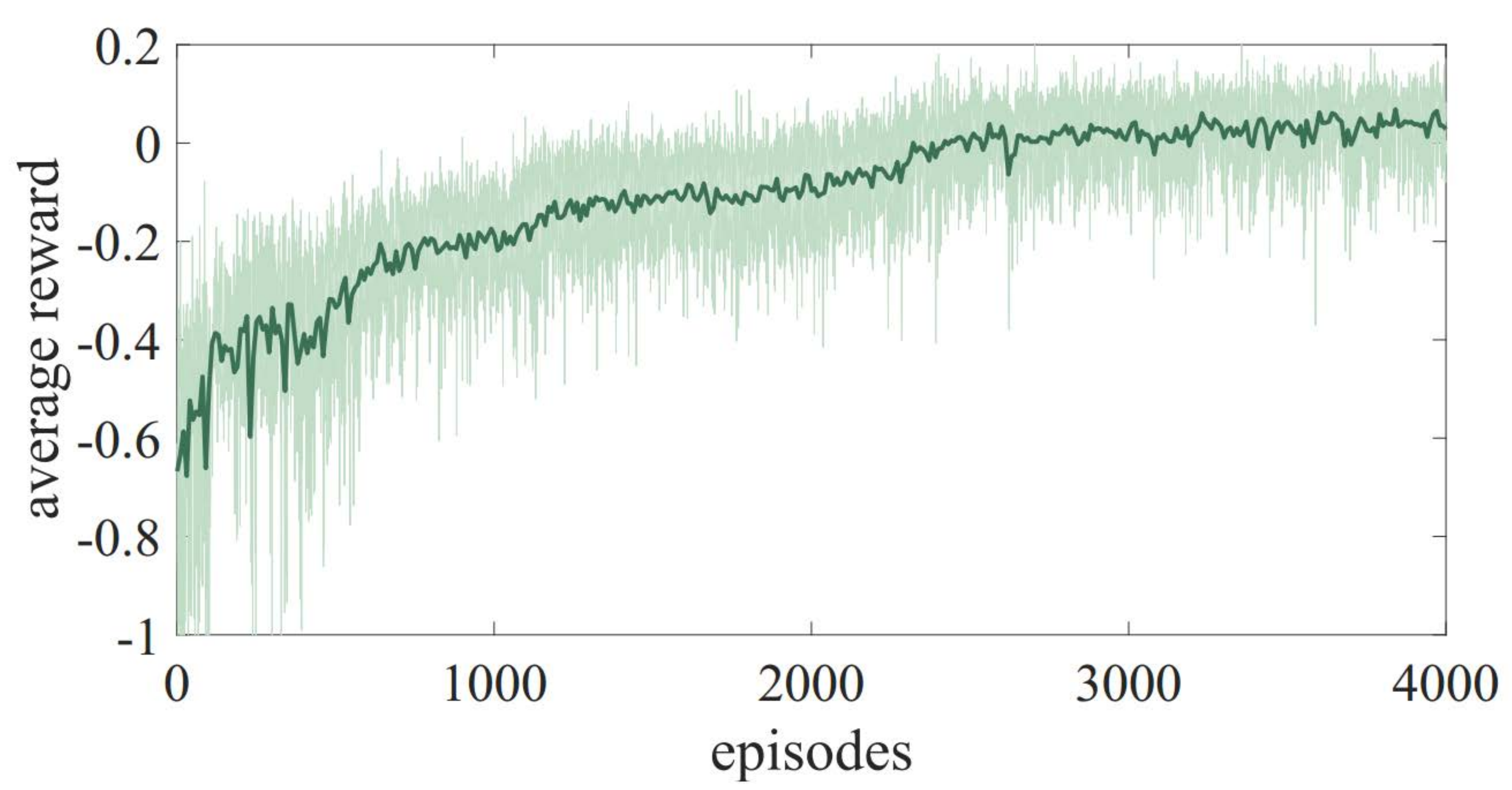}
\caption{The average reward of our units in 3 Goliaths vs. 6 Zealots micromanagement scenario during training.}
\label{fig_sim}
\end{figure}

\begin{figure}[!t]
\centering
\includegraphics[width=3.3 in, trim=0 0 0 0]{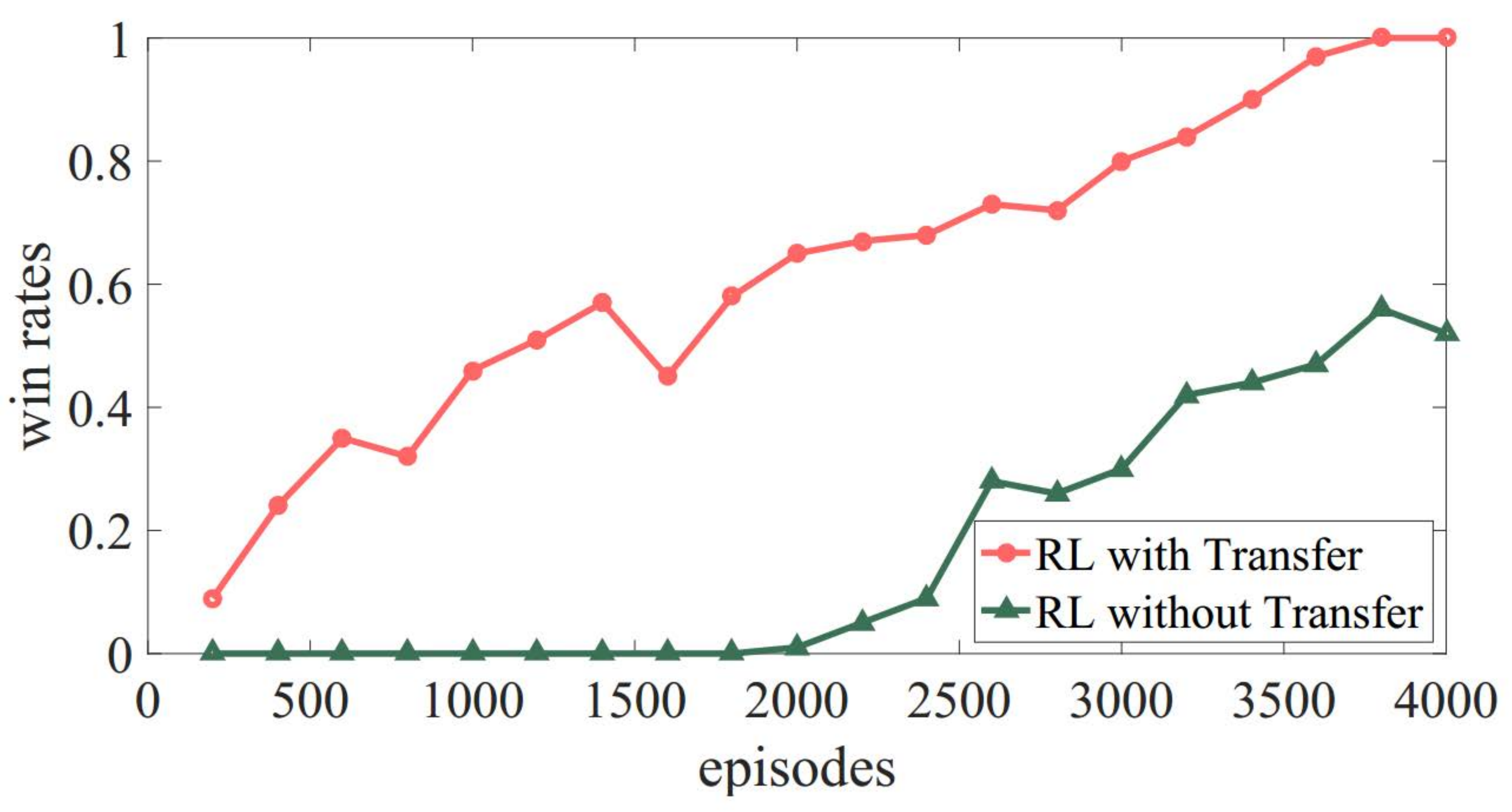}
\caption{The win rates of our units in 3 Goliaths vs. 20 Zerglings micromanagement scenario from every 200 episodes' training.}
\label{fig_sim}
\end{figure}

\begin{figure}[!t]
\centering
\includegraphics[width=3.3 in]{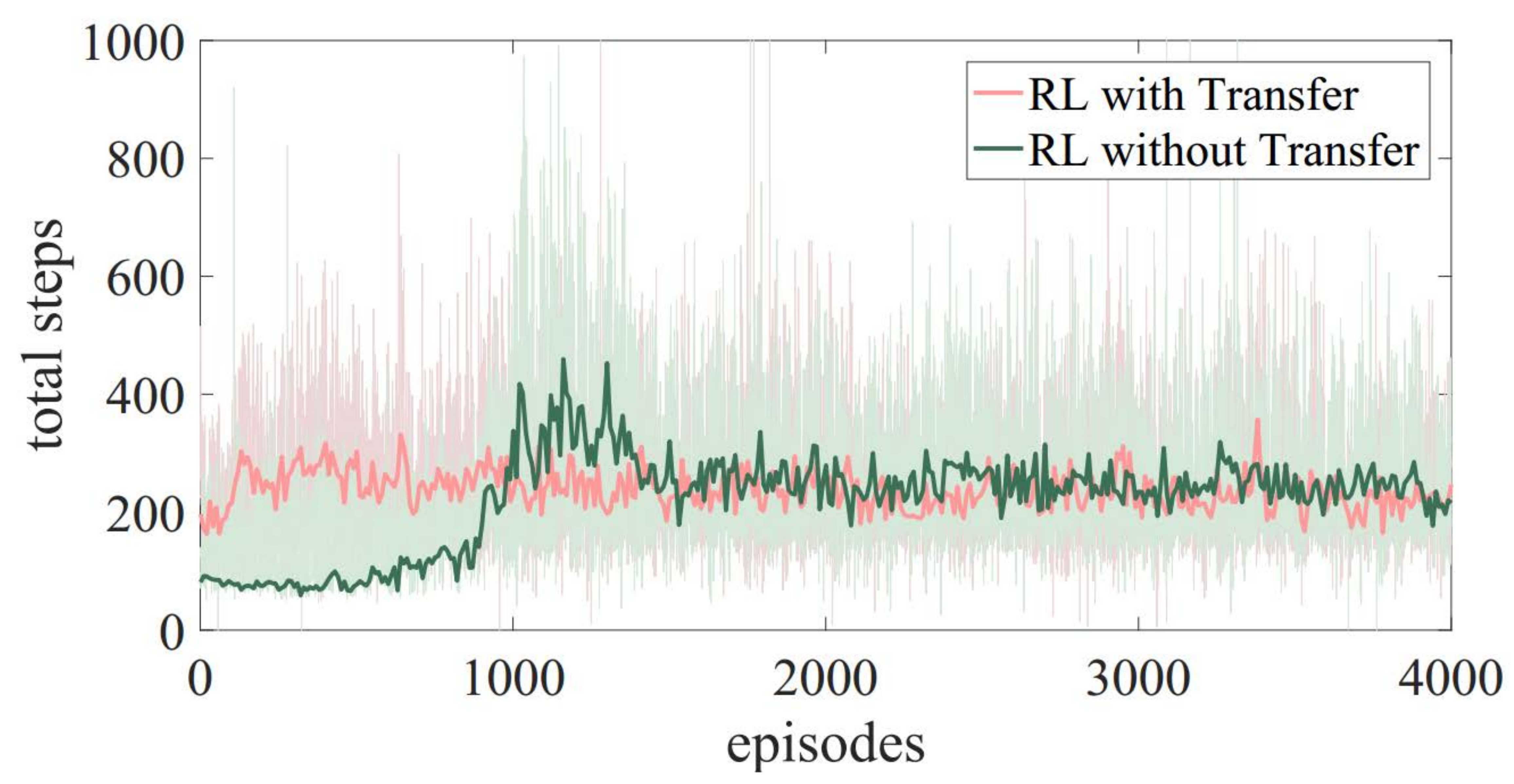}
\caption{The average episode steps of our units in 3 Goliaths vs. 20 Zerglings micromanagement scenario during training.}
\label{fig_sim}
\end{figure}

\begin{figure}[!t]
\centering
\includegraphics[width=3.3 in, trim=0 0 0 0]{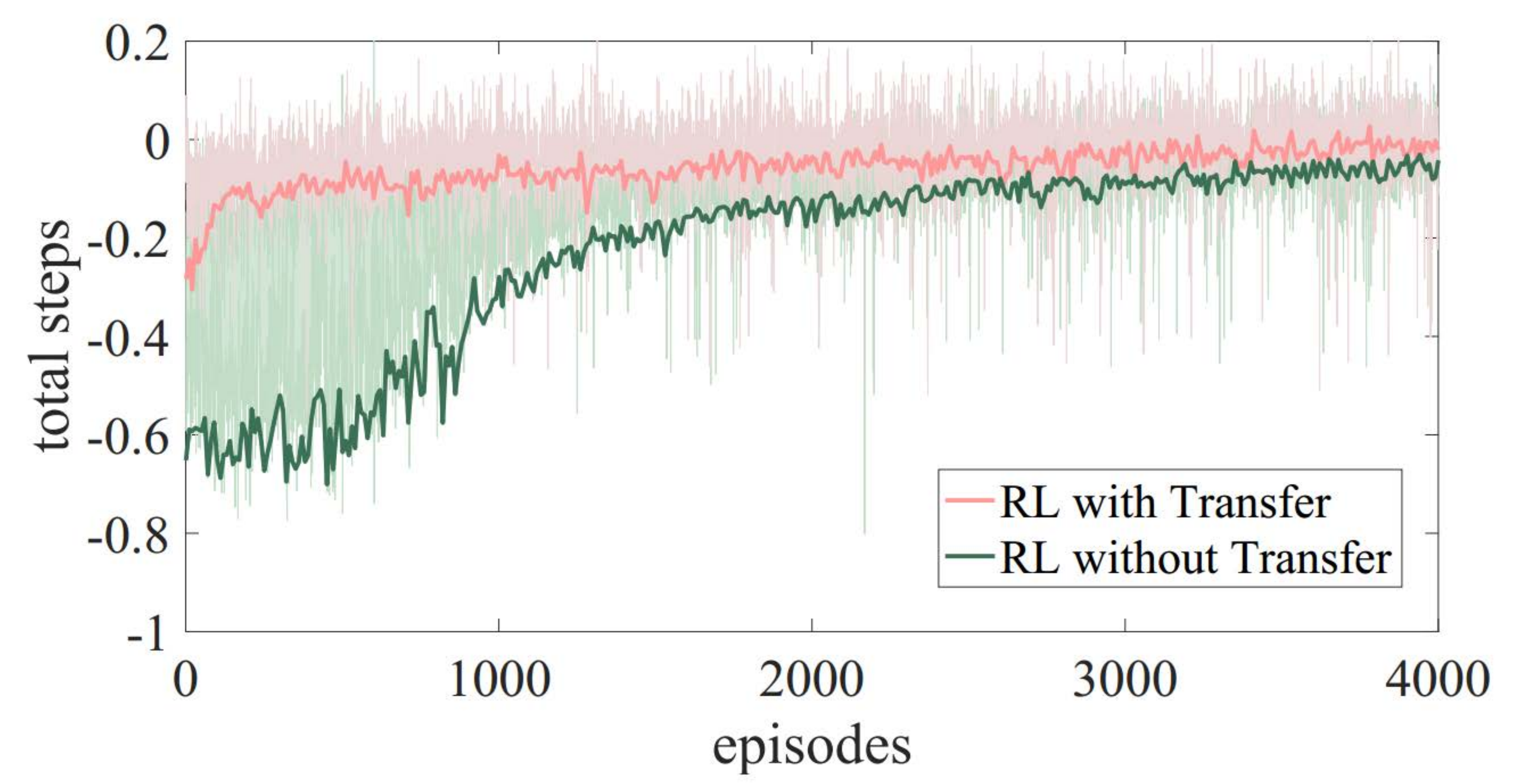}
\caption{The average reward of our units in 3 Goliaths vs. 20 Zerglings micromanagement scenario during training.}
\label{fig_sim}
\end{figure}

\subsubsection{Goliaths vs. Zerglings}
In this scenario, the enemy units are a group of Zerglings, and we reuse the well-trained model from the first scenario to initialize the policy network.
In comparison with learning from scratch, we have a better understanding of transfer learning.

\begin{itemize}
\item Win Rates: We draw the win rates in Fig. 9. When training from scratch, the learning process is extremely slow and our units can't win a game until 1800 episodes. Without transfer learning, the win rates are below 60\% after 4000 episodes. When training based on the model of the first scenario, the learning process is much faster. Even in the opening, our units win several games, and the win rates reach 100\% in the end.
\item Episode Steps: In Fig. 10, we draw the average episode steps of our three Goliaths during training. Without transfer learning, the curve has the similar trend with that in the first scenario. The average episode steps have a obvious increase in the opening and drop gradually during training. When training with transfer learning, the average episode steps keep stable in the whole training process, within the range of 200 to 400. A possible explanation is that our units have learned some basic move and attack skills from the well-trained model, and they use these skills to speed up the training process.
\item Average Rewards: We draw the average rewards of our three Goliaths in Fig. 11. When training from scratch, our units have difficulty in winning the combat in the opening and the average rewards are in a low level before 1000 episodes. The average rewards with transfer learning, by comparison, are much higher from the beginning and behave better in the whole training process.
\end{itemize}

\subsection{Large Scale Micromanagement}
In large scale micromanagement scenarios, we use curriculum transfer learning to train our Marines to play against Zerglings, and compare the results with some baseline methods.

\subsubsection{Marines vs. Zerglings}

In this section, we design a curriculum with 3 classes to train the units, as shown in Table III.
After training, we test the performance in two target scenarios: M10 vs. Z13 and M20 vs. Z30.
In addition, we use some baseline methods as a comparison, which consist of rule-based approaches and DRL approaches.
\begin{itemize}
\item Weakest: A rule-based method, attacking weakest in the fire range.
\item Closest: A rule-based method, attacking closest in the fire range.
\item GMEZO: A DRL method, based on the zero-order optimization, having impressive results over traditional RL methods\cite{Usunier2016Episodic}.
\item BicNet: A DRL method, based on the actor-critic architecture, having the best performance in most StarCraft micromanagement scenarios\cite{Peng2017Multiagent}.
\end{itemize}

In Table IV, we present the win rates of the PS-MAGDS method and baseline methods.
In each scenario, we measure our model's average win rates in 100 test games for 5 times.
In M10 vs. Z13, PS-MAGDS achieves a win rate of 97\%, which is much higher than other methods, including the recently proposed GMEZO and BicNet.
In M20 vs. Z30, PS-MAGDS has the second best performance, which is very close to the best one.

\begin{table}[!t]
\renewcommand{\arraystretch}{1.3}
\caption{Curriculum design for Marines vs. Zerglings micromanagement. M:Marine, Z:Zergling.}
\label{table_example}
\centering
\begin{tabular}{c c c c}
\hline
\hline
Scenarios & Class 1 & Class 2 & Class 3 \\
\hline
M10 vs. Z13 & M5 vs. Z6 & M8 vs. Z10 & M8 vs. Z12 \\
M20 vs. Z30 & M10 vs. Z12  & M15 vs. Z20 & M20 vs. Z25\\
\hline
\hline
\end{tabular}
\end{table}

\begin{table}[!t]
\renewcommand{\arraystretch}{1.3}
\caption{Performance comparison of our model with baseline methods in two large scale scenarios. M:Marine, Z:Zergling.}
\label{table_example}
\centering
\begin{tabular}{c c c c c c}
\hline
\hline
Scenarios & Weakest & Closest & GMEZO & BicNet & PS-MAGDS\\
\hline
M10 vs. Z13 & 23\% & 41\% & 57\% & 64\% & $\mathbf{97\%}$\\
M20 vs. Z30 & 0\%  & 87\% & 88.2\% & $\mathbf{100\%}$ & 92.4\%\\
\hline
\hline
\end{tabular}
\end{table}

\begin{table}[!t]
\renewcommand{\arraystretch}{1.3}
\caption{Win rates in various curricular scenarios and unseen scenarios. M:Marine, Z:Zergling.}
\label{table_example}
\centering
\begin{tabular}{c c c}
\hline
\hline
 Well-trained Scenarios & Test Scenarios & Win rates
\\  \hline
\multirow{6}{*}{M10 vs. Z13} & M5 vs. Z6 & 80.5\% \\
  & M8 vs. Z10 & 95\% \\
  & M8 vs. Z12 & 85\%\\
  & M10 vs. Z15 & 81\%
\\  \hline
\multirow{6}{*}{M20 vs. Z30} & M10 vs. Z12 & 99.4\%\\
  & M15 vs. Z20 & 98.2\% \\
  & M20 vs. Z25 & 99.8\% \\
  & M40 vs. Z60 & 80.5\% \\
\hline
\hline
\end{tabular}
\end{table}

We also test our well-trained models in curricular scenarios and unseen scenarios, and present the results in Table V.
We can see that PS-MAGDS has outstanding performances in these curricular scenarios.
In unseen scenarios with more units, PS-MAGDS also has acceptable results.

\subsection{Strategies Analysis}
In StarCraft micromanagement, there are different types of units with different skills and properties.
Players need to learn how to move and attack with a group of units in real time. If we design a rule-based AI to solve this problem, we have to consider a large amount of conditions, and agent's ability is also limited. Beginners of StarCraft could not win any of these combats presented in our paper. So these behaviors are highly complex and difficult to learn.
With reinforcement learning and curriculum transfer learning, our units are able to master several useful strategies in these scenarios.
In this section, we will make a brief analysis on these strategies that our units have learned.

\subsubsection{Disperse Enemies}

In small scale micromanagement scenarios, our Goliaths units have to fight against the opponent with a larger amount and more total hitpoints.
If our units stay together and fight against a group of units face-to-face, they will be destroyed quickly and lose the combat.
The appropriate strategy is dispersing enemies, and destroying them one by one.

In the first scenario, our Goliaths units have learned dispersing Zealots after training.
In the opening, our units disperse enemies into several parts and destroy it in one part first.
After that, the winning Goliath moves to other Goliaths and helps to fight against the enemies.
Finally, our units focus fire on the remaining enemies and destroy them.
For a better understanding, we choose some frames of game replay in the combat and draw units' move and attack directions in Fig. 12.
The white lines stand for the move directions and the red lines stand for the attack directions.

The similar strategy occurs in the second scenario. The opponent has much more units, and Zerglings Rush has great damage power, which is frequently used in StarCraft games. Our Goliaths units disperse Zerglings into several groups and keep a suitable distance with them. When units' weapons are in a valid cooldown state, they stop moving and attack the enemies, as shown in Fig. 13.

\subsubsection{Keep the Team}
In large scale micromanagement scenarios, each side has a mass of units.
Marines are small size ground units with low hitpoints. If they combat in several small groups, they are unable to resist the enemies.
A suitable strategy is keeping our Marines in a team, moving with the same direction and attacking the same target, as demonstrated in Fig. 14.
From these figures, we can see that our Marines have learned to move forward and retreat in a queue.

\subsubsection{Hit and Run}
Apart from the global strategies discussed above, our units have also learned some local strategies during training.
Among them, Hit and Run is the most widely used tactic in StarCraft micromanagement.
Our units rapidly learn the Hit and Run tactic in all scenarios, including the single unit's Hit and Run in Fig. 12 and Fig. 13, and a group of units' Hit and Run in Fig. 14.

\subsubsection{Existing Problems}
Although our units have learned useful strategies after training, there are still some problems in combats.
For instance, Goliaths move forward and backward now and then and don't join in combats to help other units in time.
In addition, units prefer moving to the boundary of the map, so as to avoid the enemies.

\section{Conclusion and Future Work}

\begin{figure}[!t]
\centering
\includegraphics[width=3.3 in]{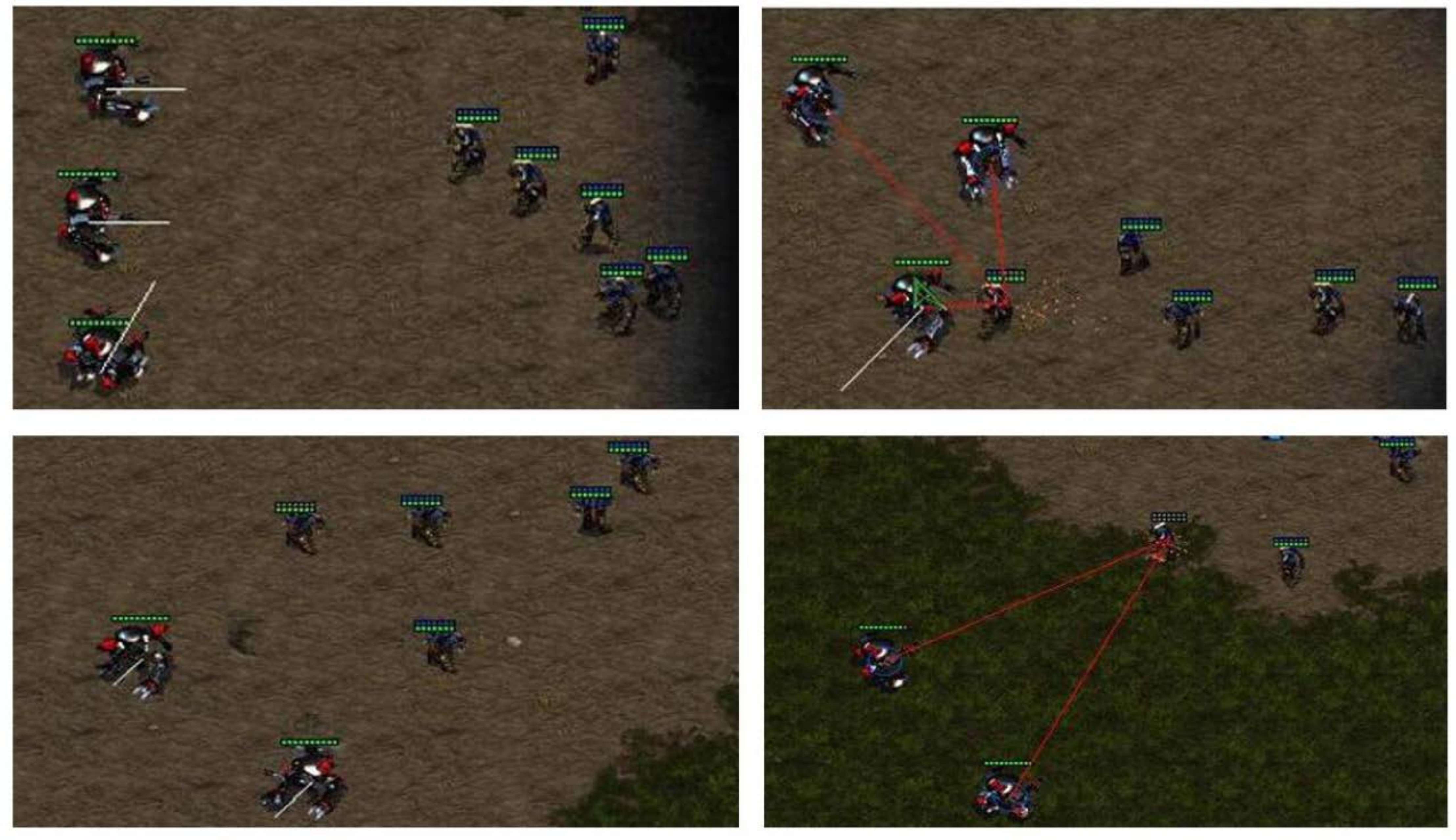}
\caption{The sample game replay in 3 Goliaths vs. 6 Zealots micromanagement scenario.} 
\label{fig_sim}
\end{figure}

\begin{figure}[!t]
\centering
\includegraphics[width=3.3 in]{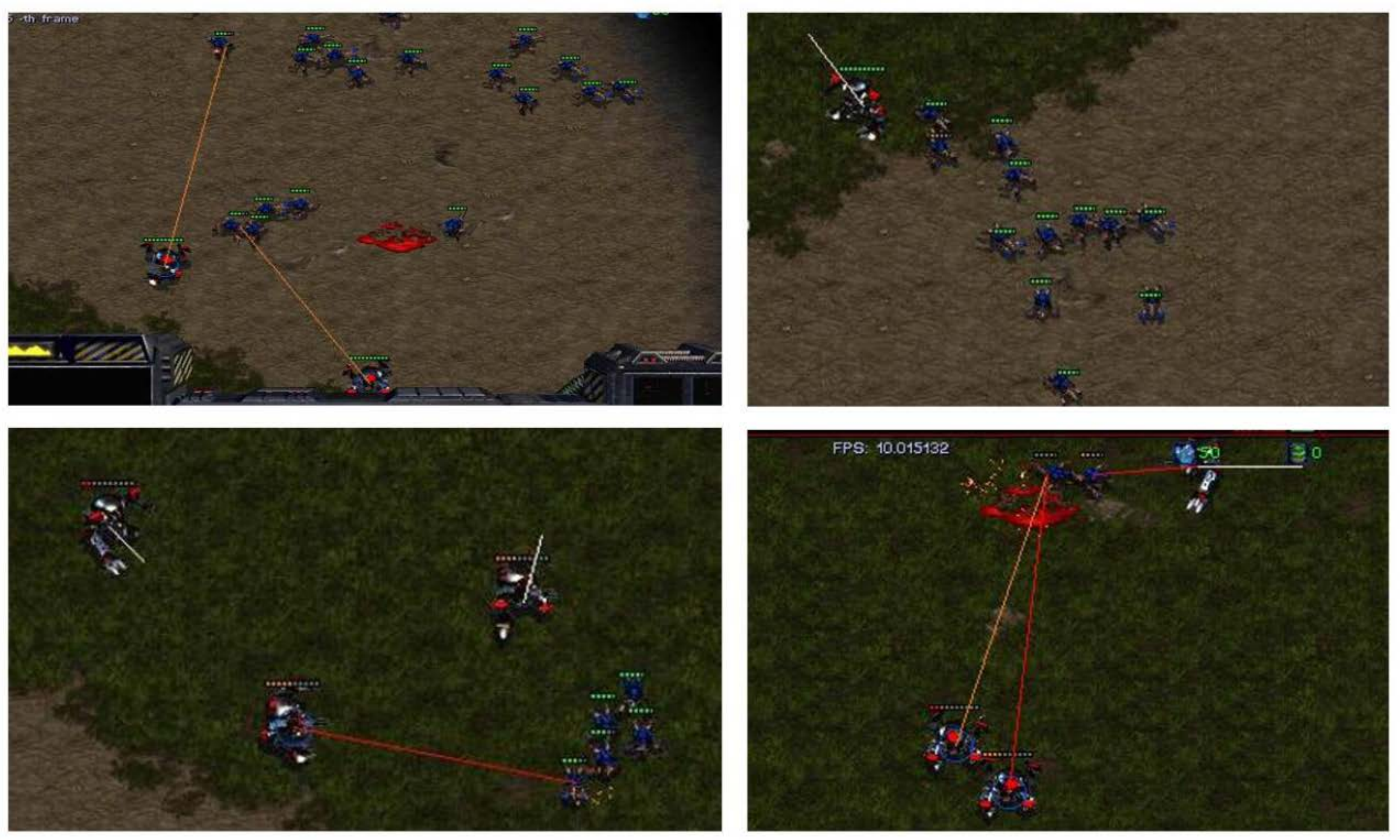}
\caption{The sample game replay in 3 Goliaths vs. 20 Zerglings micromanagement scenario.} 
\label{fig_sim}
\end{figure}

\begin{figure}[!t]
\centering
\includegraphics[width=3.3 in]{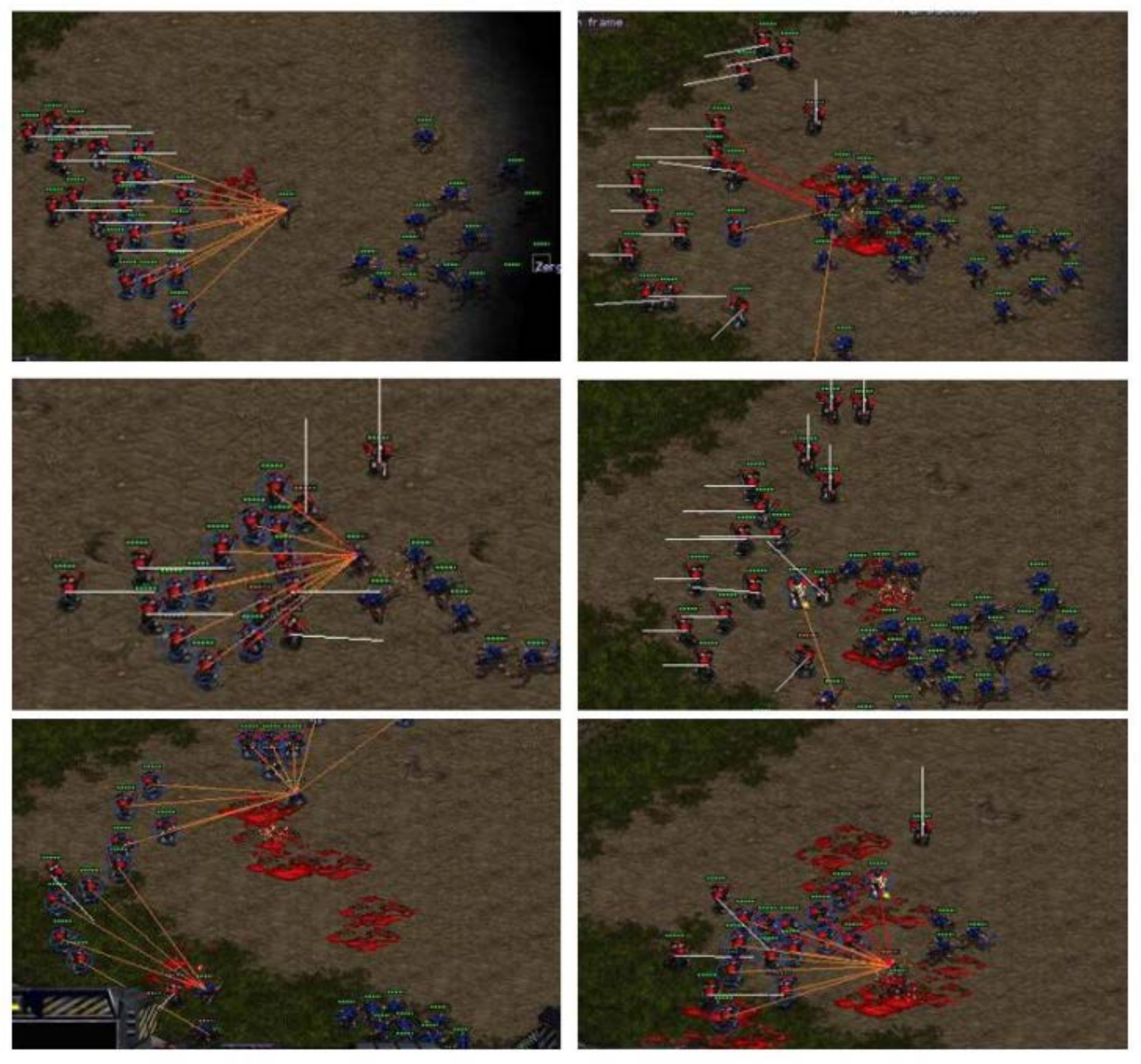}
\caption{The sample game replay in 20 Marines vs. 30 Zerglings micromanagement scenario.} 
\label{fig_sim}
\end{figure}
This paper focuses on the multiple units control in StarCraft micromanagement scenarios.
We present several contributions, including an efficient state representation,
the parameter sharing multi-agent gradient-descent Sarsa($\lambda$),
the effective reward function and the curriculum transfer learning method used to extend our model to various scenarios.
We demonstrate the effectiveness of our approach in both small scale and large scale scenarios,
and the superior performance over some baseline methods in two target scenarios.
It is remarkable that our proposed method is able to learn appropriate strategies and defeat the built-in AI in various scenarios.

In addition, there are still some areas for future work.
The cooperative behaviors of multiple units are learned by sharing the policy network, constructing an efficient state representation method including other units' information and the proposed intrinsic motivated reward function.
Although our units can successfully master some effective coordination strategies, we will explore more intelligent methods for multi-agent collaboration.
To solve the delayed reward problem in StarCraft micromanagement, we use a simple, straight and efficient reward shaping method.
Nevertheless, there are also some other methods solving the sparse and delayed rewards, such as hierarchical reinforcement learning.
Hierarchical RL integrates hierarchical action-value functions, operating at different temporal scales\cite{Kulkarni2016Hierarchical}.
Compared with the reward shaping method, hierarchical RL has the capacity to learn temporally-abstracted exploration, and gives agents more flexibility. But its framework is also much more complicated, and automatically subgoals extraction is still an open problem.
In the future, we will make an in-depth study on applying hierarchical RL to StarCraft.
At present, we can only train ranged ground units with the same type, while training melee ground units using RL methods is still an open problem.
We will improve our method for more types of units and more complex scenarios in the future.
Finally, we will also consider to use our micromanagement model in the StarCraft bot to play full the game.

\section*{Acknowledgment}
We would like to thank Qichao Zhang, Yaran Chen, Dong Li, Zhentao Tang and Nannan Li for the helpful comments and discussions about this work and paper writing.
We also thank the BWAPI and StarCraft group for their meaningful work.

\ifCLASSOPTIONcaptionsoff
  \newpage
\fi

\bibliographystyle{IEEEtran}
\bibliography{refer}


\begin{IEEEbiography}[{\includegraphics[width=1in,height=1.25in,clip,keepaspectratio]{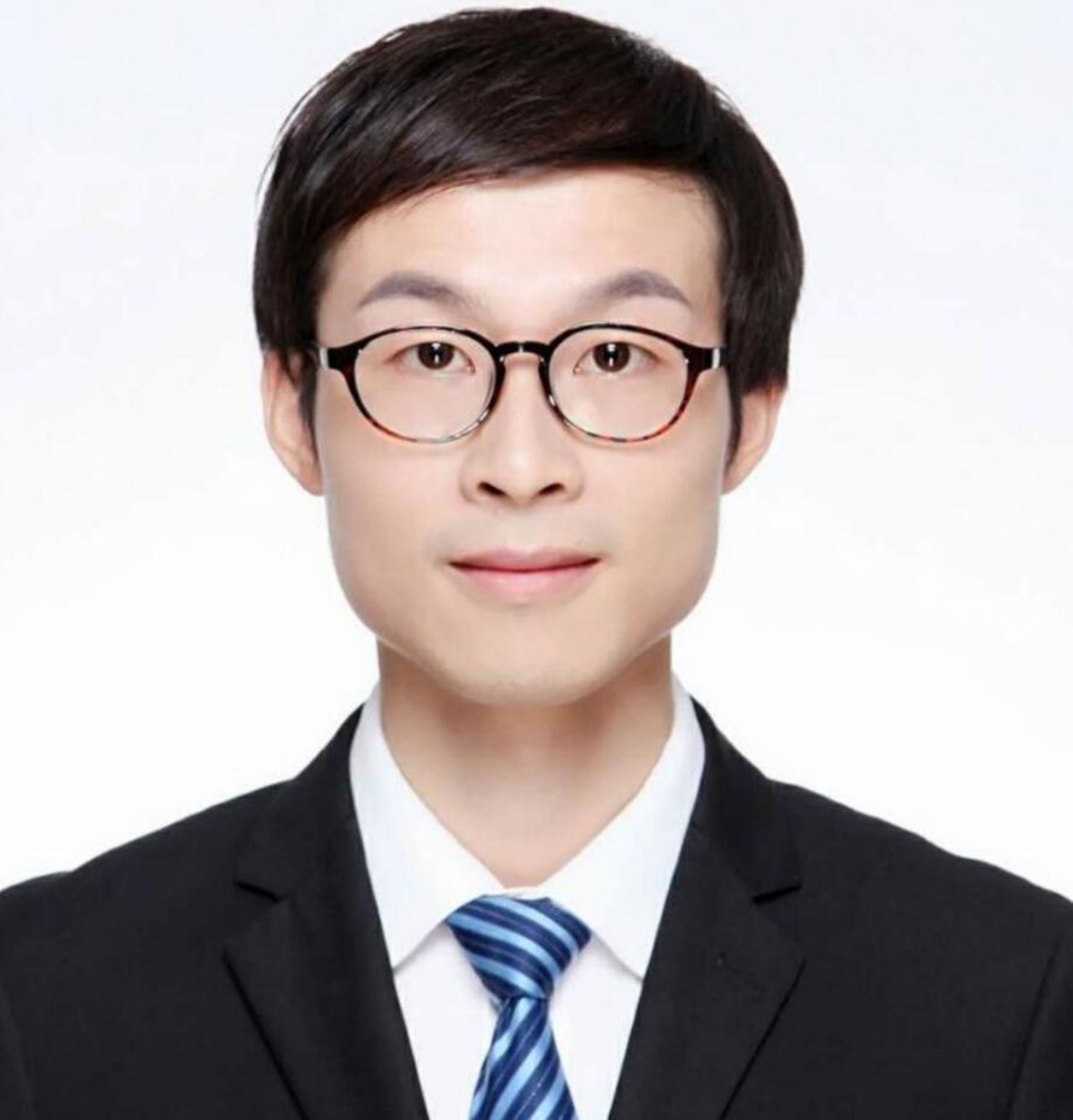}}]{Kun Shao}
received the B.S. degree in automation from Beijing Jiaotong University, Beijing, China, in 2014. He is currently pursuing the Ph.D. degree
in control theory and control engineering with the State Key Laboratory of Management and Control for Complex Systems, Institute of Automation,
Chinese Academy of Sciences, Beijing, China.
His current research interests include reinforcement learning, deep learning and game AI.
\end{IEEEbiography}

\begin{IEEEbiography}[{\includegraphics[width=1in,height=1.25in,clip,keepaspectratio]{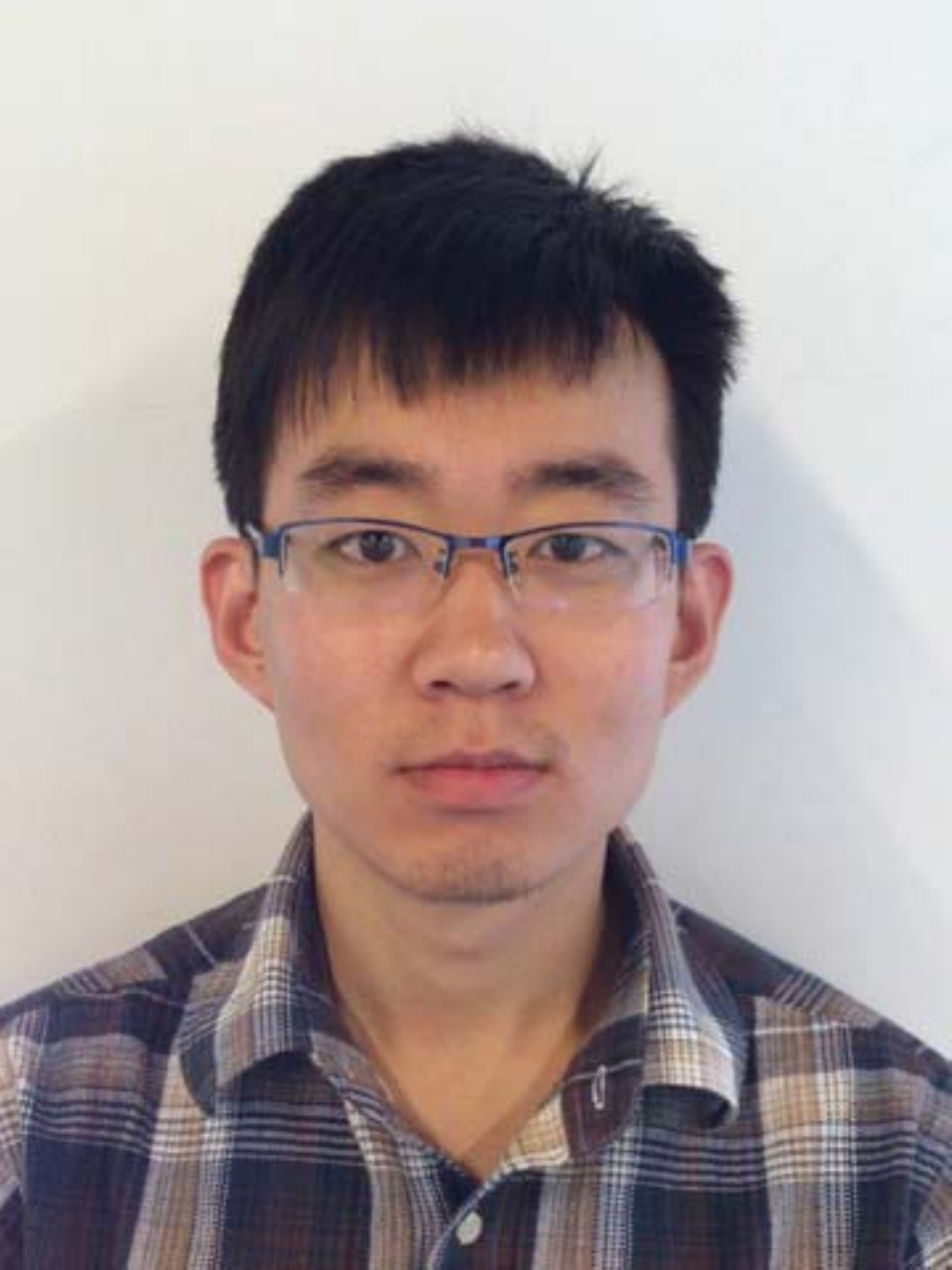}}]{Yuanheng Zhu}
received the B.S. degree from Nanjing University, Nanjing, China, in 2010, and the Ph.D. degree with the State Key Laboratory
of Management and Control for Complex Systems, Institute of Automation, Chinese Academy of Sciences, Beijing, China, in 2015.
He is currently an Associate Professor with the State Key Laboratory of Management and Control for Complex Systems, Institute of Automation,
Chinese Academy of Sciences. His research interests include optimal control, adaptive dynamic programming and reinforcement learning.
\end{IEEEbiography}

\begin{IEEEbiography}[{\includegraphics[width=1in,height=1.25in,clip,keepaspectratio]{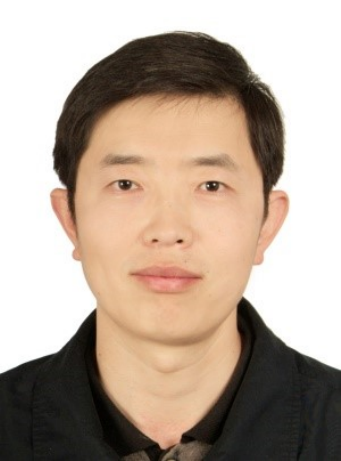}}]{Dongbin Zhao}
(M'06-SM'10) received the B.S., M.S., Ph.D. degrees from Harbin Institute of Technology, Harbin, China, in 1994, 1996, and 2000 respectively. He was a postdoctoral fellow at Tsinghua University, Beijing, China, from 2000 to 2002. He has been a professor at Institute of Automation, Chinese Academy of Sciences since 2012, and also a professor with the University of Chinese Academy of Sciences, China. From 2007 to 2008, he was also a visiting scholar at the University of Arizona. He has published 4 books, and over 60 international journal papers. His current research interests are in the area of computational intelligence, adaptive dynamic programming, deep reinforcement learning, robotics, intelligent transportation systems, and smart grids.

Dr. Zhao is the Associate Editor of IEEE Transactions on Neural Networks and Learning Systems (2012-), IEEE Computation Intelligence Magazine (2014-), etc. He is the Chair of Beijing Chapter, and was the Chair of Adaptive Dynamic Programming and Reinforcement Learning Technical Committee (2015-2016), Multimedia Subcommittee (2015-2016) of IEEE Computational Intelligence Society (CIS). He works as several guest editors of renowned international journals. He is involved in organizing several international conferences.
\end{IEEEbiography}

\end{document}